\documentclass[runningheads]{llncs}

\usepackage{eccv}

\usepackage{multirow}
\usepackage{eccvabbrv}

\usepackage{graphicx}
\usepackage{booktabs}

\usepackage[accsupp]{axessibility}  

\usepackage{my_commands} 
\usepackage{wrapfig}

\usepackage{hyperref}

\usepackage{orcidlink}

\begin{document}

\title{UMERegRobust - Universal Manifold Embedding Compatible Features for Robust Point Cloud Registration}

\titlerunning{UMERegRobust}

\author{Yuval Haitman\thanks{Corresponding Author}\orcidlink{0000-0002-6364-4028} \and
Amit Efraim\orcidlink{0000-0001-9895-6278} \and
Joseph M. Francos\orcidlink{0000-0001-9436-956X}}

\authorrunning{Y.~Haitman et al.}

\institute{Ben-Gurion University, Beer-Sheva, Israel
}

\maketitle

\begin{abstract}
In this paper, we adopt the Universal Manifold Embedding (UME) framework  for the estimation of rigid transformations and extend it, so that it can accommodate scenarios involving partial overlap and differently sampled point clouds. UME is a methodology designed for mapping observations  of the same object, related by rigid transformations, into a single low-dimensional linear subspace. This process yields a transformation-invariant representation of the observations, with its matrix form representation being covariant (\ie equivariant) with   the transformation.
We extend the UME framework by introducing a
UME-compatible feature extraction method augmented with a unique UME contrastive loss and a sampling equalizer.  These components are integrated into a comprehensive and robust registration pipeline,  named {\it UMERegRobust}. We propose the RotKITTI registration benchmark, specifically tailored to evaluate registration methods for scenarios involving large rotations.  UMERegRobust achieves better than state-of-the-art performance on the KITTI benchmark, especially when strict precision  of $(1^\circ, 10cm)$ is considered (with an average gain of $+9\%$), and notably outperform SOTA methods on the RotKITTI benchmark (with $+45\%$ gain compared the most recent SOTA method). Our code is available at \href{https://github.com/yuvalH9/UMERegRobust}{https://github.com/yuvalH9/UMERegRobust}.

\end{abstract}

\section{Introduction}
\label{sec:intro}

Point cloud registration is a critical component in many vision-based applications, such as perception for autonomous systems. The registration of point cloud observations on a rigid object, or scene, amounts to  estimating the rigid transformation relating them. However, in practical scenarios, these observations are often characterized by partial overlap as a result of being acquired from different viewpoints, as well as by different sampling patterns.

The conventional approach to addressing point cloud registration relies on three basic components: feature extraction, keypoint matching, and the estimation of rigid transformations based on corresponding keypoints.
\begin{wrapfigure}{r}{0.45\textwidth}
        \begin{minipage}[m]{1\linewidth}
        \centering\includegraphics[width=1\linewidth]{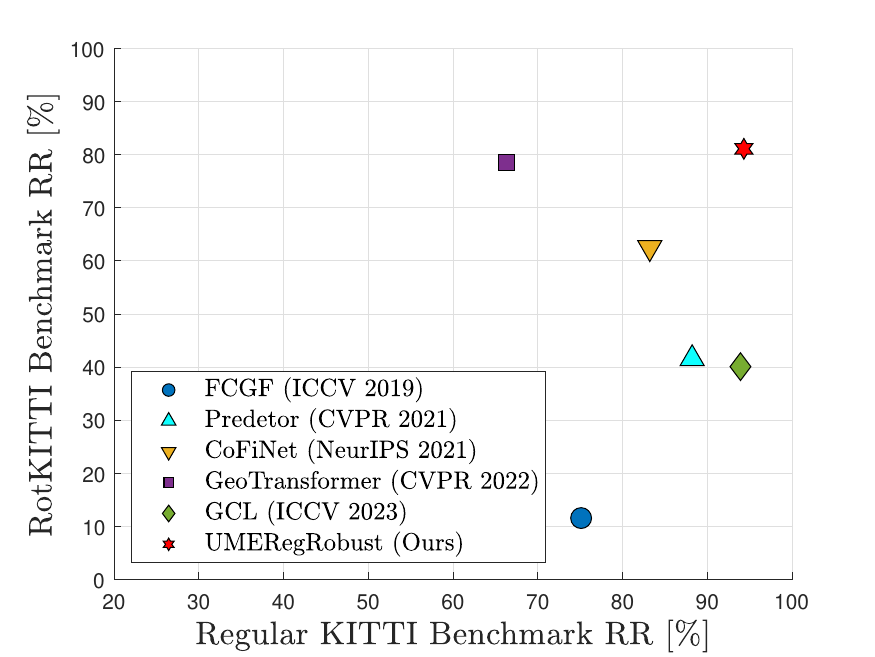}
        \end{minipage}
        \begin{minipage}[m]{1\linewidth}
        \centering\includegraphics[width=1\linewidth]{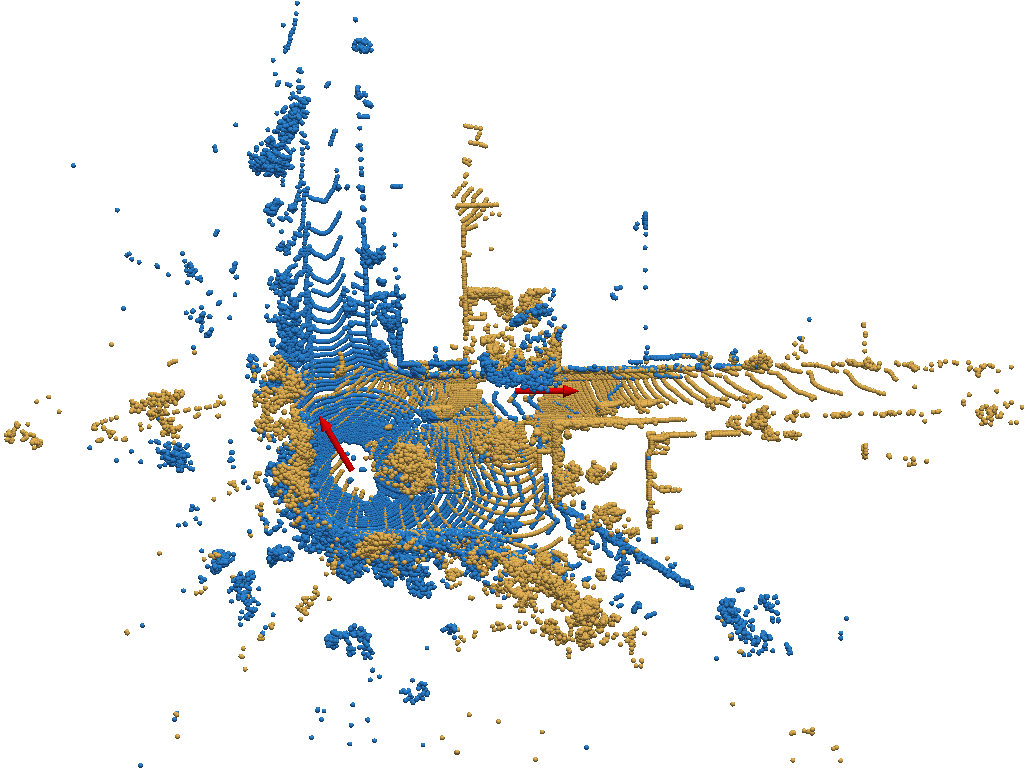}
        \end{minipage}
\caption{\textbf{Top}: Registration Recall (RR) performance of different baselines on regular KITTI ($x$-axis) and RotKITTI ($y$-axis) registration benchmarks. UMERegRobust outperforms the compared SOTA methods, on both benchmarks. \textbf{Bottom}: Registration problem from RotKITTI benchmark, highlighting significant rotation between measurements. Source and target point clouds are shown in different colors, with arrow direction representing vehicle heading, indicating a $120^\circ$ rotation problem.}
\label{fig:teaser}
\end{wrapfigure}
Recent studies have shifted towards employing learning-based  feature extraction, demonstrating promising results even in scenarios with significant variations among observations \cite{FCGF,3DFeatNet, overlapPredator, GeoTransformer, GCL}. Despite the notable advancements facilitated by learning-based techniques, the core elements of the registration pipeline have remained largely unchanged, relying heavily on traditional methods such as  Horn \cite{Horn87} for estimating rigid transformations.

In this study, we adopt the Universal Manifold Embedding (UME) framework \cite{UME,RTUME_JMIV} for the estimation of rigid transformations and extend it, so that it can accommodate scenarios involving large transformations, partial overlap and differently sampled point clouds. UME is a methodology designed for mapping observations (\eg, images, 3D point clouds, etc.) of the same object, related by rigid transformations, into a single low-dimensional linear subspace. This process yields a transformation-invariant representation of the observations, with its matrix form representation being covariant (\ie, equivariant) with  the transformation. This duality is advantageous, as the invariant representation facilitates matching of corresponding observations, while the covariant property of the matrix representation enables estimating the transformation relating the observations. A prerequisite for generating a UME descriptor for a given observation is to define a transformation-invariant function over the observations, which we name the observation coloring function.

Unlike registration methods that rely solely on point-wise matched point's coordinates information for transformation estimation \cite{Horn87}, the UME leverages both local geometric correspondences and their matched neighborhood coloring. This approach leads to a closed-form solution that offers greater accuracy and robustness.

While the UME registration method has demonstrated effectiveness on synthetic closed objects like \cite{MODELNET40, shapenet},  it encountered difficulties with outdoor and indoor scans  due to lack of invariant coloring functions, robust to sampling variations and partial overlap. In this study, we propose a novel UME-compatible coloring solution that is both invariant to transformations and robust to sampling variations. We introduce a coloring module based on a Fully-Convolutional neural network, which we train using a novel UME contrastive learning approach on pairs of point cloud observations, pre-processed by a  Sampling-Equalizer Module aimed at enhancing robustness to sampling variations. Finally, we  introduce a comprehensive registration pipeline built upon the UME framework.

We showcase the performance of our proposed method on various registration benchmarks including outdoor(KITTI \cite{KITTI}, nuScenes \cite{nuscenes2019}) and indoor (3DMatch \cite{3DMatch}). For many perception tasks, especially in autonomous driving, high precision is crucial for system safety and performance. Hence, we evaluate and compare our method on the outdoor benchmarks with a strict precision criteria of $(1^\circ, 10cm)$. We also suggest RotKITTI - a new outdoor registration benchmark focusing on problems with large rotations. This type of problems is of high importance in evaluating registration methods for SLAM systems loop closure \cite{dellenbach2022ct}. We compare the performance of the proposed method against a large set of baseline methods (some of the comparisons are shown in Fig. \ref{fig:teaser}).

The main contributions of this paper are:
\begin{enumerate}
     \item We introduce a novel Universal Manifold Embedding (UME) compatible coloring method, augmented with a unique UME contrastive loss and a Sampling Equalizer Module. The proposed UME-compatible coloring provides  an {\bf enabler} that facilitates high performance UME-based registration for general scenes such as outdoor/indoor scans. Additionally, we present a comprehensive and robust RANSAC-Free registration pipeline for 3D point clouds, comprising a dedicated feature extractor for UME descriptor generation, a matched manifold detector for point cloud matching, a UME-based estimator for hypothesis estimation, and a hypothesis selection module for selecting the best estimator.

     \item We propose the RotKITTI and RotnuScenes registration benchmarks, specifically tailored to evaluate registration methods for scenarios involving large rotations. These benchmarks are crucial for assessing methods intended for integration into SLAM systems, particularly for the task of loop closure.

     \item We achieve better than state-of-the-art performance on the KITTI benchmark, especially when strict precision  of $(1^\circ, 10cm)$ is considered (with an average gain of $+9\%$), and notably outperform SOTA methods on the RotKITTI benchmark (with $+45\%$ gain compared the most recent SOTA method).
\end{enumerate}

\section{Related Work}
\label{sec:related_work}
When the relative pose of point clouds undergoing registration is unknown, the most common approach begins with matching key points, followed by estimating transformations between these correspondences using Horn's solution via constrained least squares estimation \cite{Horn87}. Due to the presence of outliers in the estimated correspondences, robust registration algorithms, such as Random Sample Consensus (RANSAC), are necessary to estimate the registration parameters and achieve an approximate alignment \cite{RANSAC}. Following this initial alignment, local optimization is typically performed (e.g., \cite{ICP, NDT}).

Recent advancements in point cloud registration have primarily focused on developing high-performance feature descriptors, while the core registration methodology has remained largely unchanged. Modern 3D feature descriptors incorporate FCGF \cite{FCGF} and KPConv \cite{KPConv} architectures as part of broader solutions for both indoor and outdoor registration challenges. Overlap Predator \cite{overlapPredator}, for example, employs KPConv convolutions with self and cross-overlap attention modules to learn local and co-contextual information. GeoTransformer \cite{GeoTransformer} uses similar backbone with a geometric transformer module in a hierarchical approach to estimate point correspondences. GCL \cite{GCL} adopts a unique group-wise contrastive learning approach, achieving high performance in scenarios with very low overlap and large translations. CofiNet \cite{cofinet} combines a coarse-to-fine matching approach with attentional feature aggregation to generate point correspondences.

Other methods focus on indoor scenarios only. Lepard \cite{lepard2021}, for instance, utilizes  positional encoding along with a reposition technique to modify cross-point cloud relative positions, applicable for both rigid and non-rigid transformation estimation. PEAL \cite{peal2023} initially employs a GeoTransformer model to estimate the overlap region between two-point clouds and then uses a similar model to estimate the transformation based on this overlap. YOHO \cite{yoho} and RoReg \cite{roreg} propose rotation-invariant descriptors by averaging descriptors obtained over multiple rotations, addressing rotation invariance but not translation. E2E \cite{e2e_iros2023} suggests a rotation-invariant descriptor using Spherical CNN \cite{spcnn}.

A common aspect of these approaches is that registration is achieved by aligning the estimated set of corresponding points, meaning only the coordinates of the matched points are used in the transformation estimation \cite{Horn87}. In contrast, the proposed UME registration pipeline uses both the estimated correspondences and their extracted features, thereby embedding valuable contextual information in the transformation estimation process.

\section{Method}
\label{sec:method}
\begin{figure}[tb]
  \centering
  \includegraphics[width=\linewidth]{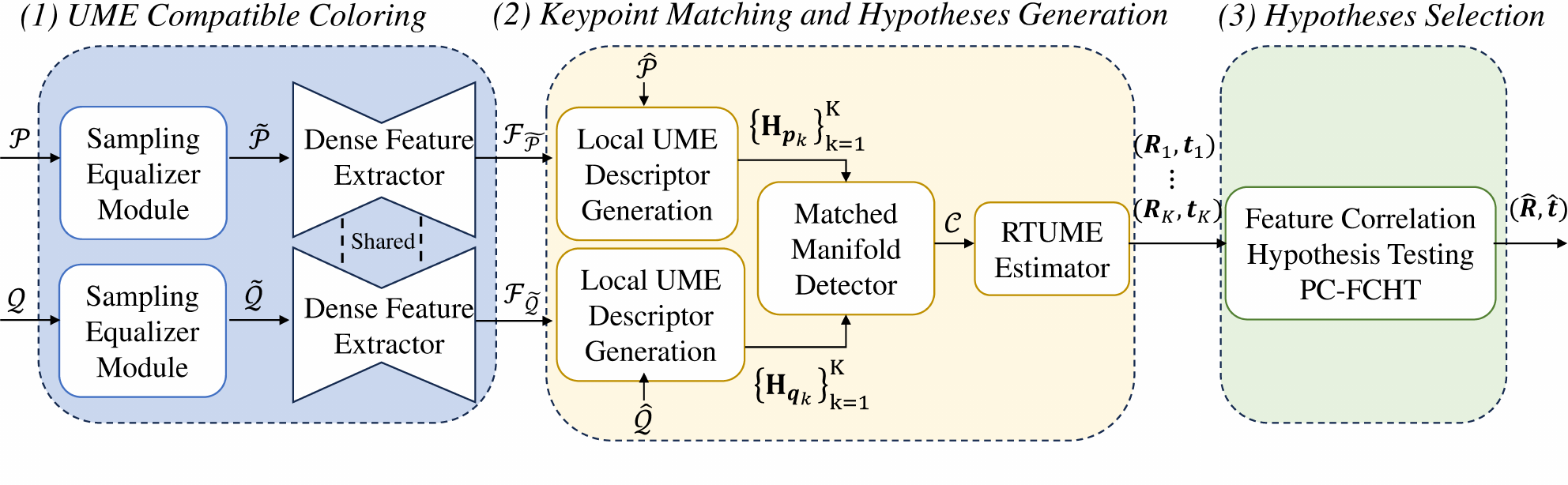}
  \caption{\textbf{UMERegRobust Overview.} $\cP, \cQ$ are the input point clouds; (1) The UME Compatible Coloring Module resamples the point clouds into a uniform grid generating $\widetilde{\cP}$ and $\widetilde{\cQ}$, then assigns  each point with a transformation invariant feature vector creating the colored point clouds $\cF_{\widetilde{\cP}}, \cF_{\widetilde{\cQ}}$. (2) Local UME descriptors, generated on a down-sampled versions of the point clouds $\widehat{\cP}, \widehat{\cQ}$, are denoted by $\{\vH_{\bsp}\}_{k=1}^K,\{\vH_{\bsq}\}_{k=1}^K$, respectively. A Matched Manifold Detector identifies corresponding local UME descriptors, forming a set of $K$ putative matched pairs $\cC$. For each matched pair, an estimated transformation is obtained using the RTUME estimator. (3) Feature correlation is used to select the hypnosis that maximize the feature correlation between the point clouds.}
  \label{fig:UMERegRobust-pipeline}
\end{figure}

Let $\cP,\cQ\subset \R^3$ be two partially overlapping point clouds related by a rigid transformation and sampling variations. We  solve the registration problem between $\cP$ and $\cQ$ by estimating the rigid transformation $(\bsR,\bst)$ where $\bsR\in \SO(3)$ and $\bst\in\R^3$.
The \emph{UMERegRobust} pipeline is depicted in Fig.\ref{fig:UMERegRobust-pipeline}. It incorporates three primary components. (1) UME Compatible Feature Extractor which assigns dense features to the input point clouds. It is designed to satisfy the requirements of the UME framework; (2) Key-point matching and hypothesis generation module, that employs the UME descriptors evaluated from the point cloud dense features, to initially establish putative matches through a Matched Manifold Detector (MMD), followed by generation of hypothesized estimates of the rigid transformation using UME-based estimators; (3) Hypothesis selection module, that selects the best transformation estimate through maximization of the point clouds feature correlation.

\subsection {Universal Manifold Embedding Overview}
\label{sec:ume_overview}

Let $s$ be a 3D object and $\cO_s\subset \R^3$ be the set of all possible  observations on $s$ generated by the action of the transformation group $\SE(3)$, \ie,  $\cO_s$ is the \emph{orbit} of $s$. It has been shown \cite{RTUME_JMIV} that the Universal Manifold Embedding (UME) provides a mapping from the orbit of possible observations on the object, generated by the action of the transformation group $\SE(3)$ to a single low dimensional linear subspace of Euclidean space. This linear subspace is \emph{invariant} to geometric transformation and hence is a unique representative of the object, regardless of its observed pose, while its matrix representation is \emph{covariant} (\ie equivariant) with the transformation. It thus naturally serves as an invariant statistic for solving problems of joint detection and transformation estimation.

Let $o\in \cO_s$ be a point cloud observation on the object $s$ and $f:\R^3\to\R^d$ is a function that assigned a real valued vector to each point in the observation. We name $f$ the \emph{observation coloring function} and $f_o(\vx)=\{f(\vx) | \vx \in o\}$ the \emph{colored observation}. The UME matrix of the colored observation $h(\vx)=f_o(\vx)$, which serves as  the \emph{UME Descriptor} of the observation is given by:

\begin{equation} \label{eq:TMapping}
\vT(h)=
\left[ {{\begin{array}{*{20}c}
{\int\limits_{ \R^3 }  w_1 \circ h(\vx)} d\vx &   {\int\limits_{ \R^3 } x_1 w_1 \circ h(\vx)} d\vx &  \hdots   &\; \int\limits_{ \R^3 } x_3 w_1 \circ h(\vx)d\vx  \\
& \vdots & \\
\int\limits_{ \R^3 } w_{M} \circ h(\vx)d\vx & {\int\limits_{ \R^3 }} x_1 w_{M} \circ h(\vx) d\vx &   \hdots   & {\int\limits_{ \R^3 } x_3 w_{M}\circ h(\vx)d\vx}\\
 \end{array} }} \right]
\end{equation}
where $\{w_m\}_{m=1}^M$ are measurable functions aimed at generating many compandings of the observation and $\vx=[x_1, x_2, x_3]^T$. A necessary condition for the applicability of the UME representation for point cloud registration is that the coloring function $f$  be invariant to the transformation \ie,that  $f(\vx) = f(\bsR \vx + \bst)$ where $(\bsR,\bst)\in\SE(3)$.
For any two colored observations $h(\vx)=f_{o}(\vx)$ and $g(\vx)=f_{o'}(\vy)$ where $o,o'$ are related by rigid transformation, their corresponding UME matrices are related by
$\vT(h)= \vT(g) \vD^{-1}(\bsR,\bst), \quad \vD(\bsR,\bst)= \begin{bmatrix} 1 & \bst^T \\ \bf 0 & \bsR^T \end{bmatrix}
$ 
Therefore, the UME matrix representation of the observation $\vT(h) $ is covariant with the transformation, where the linear subspace spanned by its column space $\langle \vT(h) \rangle$ is invariant to the transformation (since $\vD$ is an inevitable matrix).
Note that $\langle \vT(h) \rangle \in \Gr(M,4)$, where  $\Gr(M,4)$  is the Grassmann manifold of $4$-dimensional linear subspaces of an  $M$-dimensional Euclidean space.

\subsubsection{UME Local Descriptor.}
Since the UME representation  in \eqref{eq:TMapping} was originally developed for an observation on a single object, adaptations to the general case where multiple objects are present (\eg, LiDAR scans), are required.  Thus, rather than generating a single UME descriptor for the entire observation, multiple local UME descriptors are generated. Given that  the entire observation undergoes the same rigid transformation, each local descriptor undergoes the same transformation: Let $h$ be a colored observation of the point cloud $\cP\subset\R^3$ (colored by the observation coloring function) and $\{\bsp_k\}_{k=1}^K\subseteq\cP$ a set of selected points each with corresponding  local neighborhood of radius $R$, $\cP_k=\{\bsp_i\in \cP \big| \|\bsp_i-\bsp_k\|_2 \leq R\}$. The \emph{Local UME Descriptor} of $\bsp_k$ denoted by $\vH_{\bsp_k}$ is obtained using a local adaption of \eqref{eq:TMapping} where the integrals are evaluated locally on the subset $\cP_k$, and thus for correctly matched points in two point clouds to be registered, the UME relations holds.

\subsection{UME Compatible Features}\label{sec:UME_coloring}
As outlined in Sec. \ref{sec:ume_overview}, a necessary step in adapting the UME framework for point cloud observations is to define a coloring function that assigns each point in the cloud with a value. This function should be invariant to the action of rigid transformations. For closed objects, natural candidates for the coloring function, such as distance from the center of mass, or surface curvature have been employed \cite{RTUME_JMIV}. However, these functions are  not optimized for  scans of outdoor or indoor scenarios.
Moreover, since the point clouds to be registered may be acquired by different sensors, at different times and from different points of view, the coloring function has to be robust to different sampling patterns where it is evident that the aforementioned candidates may be sensitive to them.

Since defining such a function analytically can be very hard, if at all possible, we adopt a data-driven approach for implementing the coloring function so that it is compatible with the UME framework such that three primary requirements are satisfied: (1) Invariance to rigid transformations; (2) Robustness to sampling variations; (3) High expressibility of the observation. The UME compatible coloring module, illustrated in Fig.\ref{fig:UMERegRobust-pipeline}, comprises two main building blocks. To address sampling differences, we first introduce a Sampling Equalizer Module, serving as a preprocessing technique for the subsequent dense feature extraction, implemented by a deep neural network.

\subsubsection{Sampling Equalizer Module.}
To mitigate the mismatch in the sampling patterns of a pair of point cloud observations we implemented a two-stage \emph{Sampling Equalizer Module} (SEM). Initially, we employ an off-the-shelf surface reconstruction technique for point cloud observations. Numerous options exist in the literature, ranging from classical methods such as Poisson reconstruction \cite{kazhdan2006poisson} to DNN-based approaches. In this study, we adopted the Neural Kernel Surface Reconstruction (NKSR) \cite{NKSR}, a single-shot surface reconstruction technique, chosen for its robust and real-time performance compared to other methods. Next, the reconstructed surface generated from the point cloud observation is sampled into a uniform grid with voxel size of $\rho$, resulting in a new, uniformly sampled point cloud denoted in Fig. \ref{fig:UMERegRobust-pipeline} by $\widetilde{\cP}, \widetilde{\cQ}$. In the Supplementary, examples are provided of point cloud observations - before and after the SEM. The application of SEM contributes to UME compatibility in two main ways. First, we can control the density of the processed point clouds by adjusting the value of $\rho$, thereby enhancing the precision of numerically evaluating the integrals in \eqref{eq:TMapping}. Second, by introducing uniformity in sampling the point clouds, we reduce mismatches  between local UME descriptors of matching neighborhoods, that may result from evaluating the integrals in \eqref{eq:TMapping} over differently sampled neighborhoods.

\subsubsection{Dense Feature Extractor.}
We adopt a fully convolutional neural network as our feature extractor, employing a similar architecture to \cite{FCGF}. Our feature extractor is a Unet-shaped DNN utilizing sparse 3D convolutions with skip-connections and it is the only learnable module in the entire registration pipeline (see detailed implementation in the Supplementary). The feature extractor assigns a feature vector to every point in the input point cloud. These features encapsulate both global and local contextual information based solely on the observation geometry, thus creating a coloring function with high expressibility of the observation.
Thus, the  coloring function is the result of two cascaded blocks: the SEM and the dense feature extractor. Invariance to rigid transformations is achieved by training the model using augmentations, along with utilizing UME-compatible losses that optimize the coloring module to match the UME theoretical requirements.

\subsection{Keypoint Matching and Hypothesis Generation}\label{sec:hyp_gen}
\subsubsection{Matched Manifold Detector.}
Since the linear subspace spanned by each UME local descriptor is invariant to rigid transformations, UME local descriptors of matching points  are mapped into the same point on the Grassmann Manifold $\Gr(M,4)$, regardless of their pose. Therefore, an effective approach to match these descriptors is by assessing their affinity on the manifold. To select the putative points to be matched, we create a down-sampled version of the input point clouds $\widehat{\cP}=\{\bsp_k\}_{k=1}^K\subseteq \cP$, $\widehat{\cQ}=\{\bsq_k\}_{k=1}^K\subseteq \cQ$. For each selected point we generate its local UME descriptor from its corresponding local neighbourhood $\{\cP_k\}_{k=1}^K$,$\{\cQ_k\}_{k=1}^K$ using the assigned features $\{\cF_{\cP_k}\}_{k=1}^K$, $\{\cF_{\cQ_k}\}_{k=1}^K$ to obtain $\{\vH_{\bsp_k}\}_{k=1}^K$, $\{\vH_{\bsq_k}\}_{k=1}^K$, respectively. Let $\langle \vX \rangle$ and $\langle \vY \rangle$ be two linear subspaces spanned by the columns of full rank matrices $\vX,\vY\in\R^{M\times r}$, respectively. Their distance on $\Gr(M,r)$ is \cite{Edelman}:
\begin{equation} \label{eq:dpf}
d_{pF}\left(\vX,\vY\right) = \frac{1}{\sqrt{2}}\|\vP_{\vX} -\vP_{\vY} \|_F = \|\sin(\bstheta)\|_2
\end{equation}
where $\vP_\vX,\vP_\vY$ are the orthogonal projection matrices on the column space of $\vX$ and $\vY$, respectively. The distance is also equivalent to the $\ell_2$-norm of the sine vector of the principal angles between the two subspaces where the maximal distance is $\sqrt{r}$, and is zero for identical subspaces.
Matched Manifold Detection (MMD) is defined as the operation of detecting an observation that belongs to the same orbit (as defined in Sec. \ref{sec:ume_overview}) as a given query observation, regardless of their relative transformation. Ideally,  MMD operation is equivalent to finding a UME descriptor with zero $d_{pF}$ distance to the tested query.
\begin{wrapfigure}{r}{0.43\textwidth}
  \centering
  \includegraphics[width=0.43\textwidth]{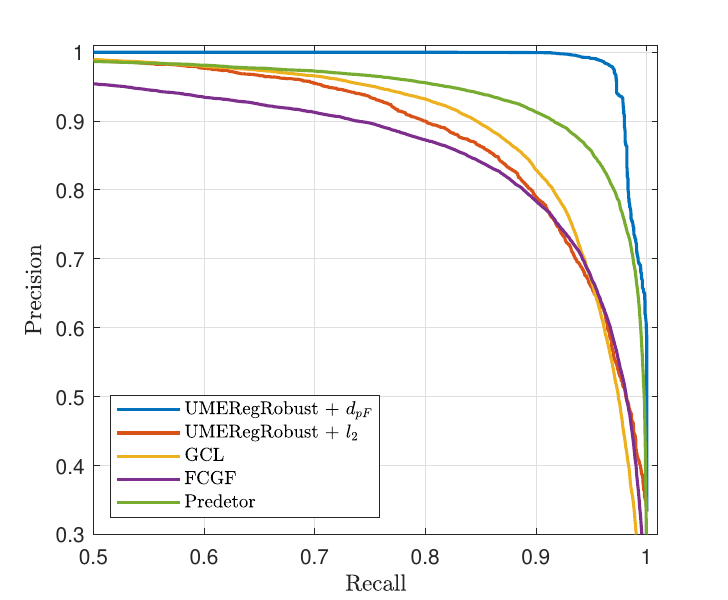} %
  \caption{PR-Curve of Keypoint matching performance under rigid transformation, partial overlap and sampling variations of UME Local descriptor vs. other descriptors.}
  \label{fig:mmd_pr}
\end{wrapfigure}
However, in practice, local UME descriptors of matching neighborhoods are not identical due to the differences in the sampling of these neighbourhoods. Hence, the zero $d_{pF}$ constraint is relaxed to a minimal distance, and the keypoint matching in the evaluation step is done by employing the Hungarian Algorithm \cite{kuhn1955hungarian} on the pair-wise $d_{pF}$ distance matrix. Note that while the standard point-to-point matching \cite{FCGF,cofinet, GeoTransformer, D3Feat, overlapPredator} may be sensitive to the noisy nature of the measurements, applying neighbourhood-to-neighbourhood matching using a Matched Manifold Detector applied to the UME local descriptors provides higher robustness.

To demonstrate the robustness of local UME descriptors in keypoint matching under rigid transformations, partial overlap, and varying sampling patterns, we compared their performance to that of STOA local point-wise descriptors. Fig. \ref{fig:mmd_pr} presents a precision-recall curve  comparing local UME descriptors (distances evaluated by $d_{pF}$) matching accuracy, to other descriptors matching accuracy. The results clearly show the superior accuracy and robustness of local UME descriptors, especially at Recall $@0.95$, where they achieve a precision gain of approximately $15\%$ over the leading competitor. Additionally, we evaluated the matching performance of Local UME point-wise descriptors using the $\ell_2$ metric, which resulted in a significant deterioration of $30\%$ compared to the UME neighborhood-based descriptor and $d_{pF}$ distance. These findings underscore the significant advantage of using UME-based distance between neighborhoods over other local point-wise descriptors.

\subsubsection{Hypothesis Generation.}
Following the detection of matching neighborhoods, the Rigid Transformation UME (RTUME) estimator \cite{RTUME_JMIV} is employed for generating multiple hypotheses of the underlying transformation relating every pair of matching neighborhoods.
More specifically, for every one of  $K$ putative matched pairs,
the RTUME estimator generates an hypothesized estimate, with total of $K$ hypotheses $\{(\bsR_k,\bst_k)\}_{k=1}^K$.
Note that unlike common estimation methods, \eg, \cite{Horn87} that require multiple point correspondences to generate a {\it single} estimate of the transformation, the RTUME provides a transformation estimate from every single pair of matched neighborhood descriptors.
The RTUME estimate of the transformation employs information from the entire neighborhood of the matched point thus leveraging the covariant  property of the UME descriptor. Unlike competing methods \cite{cofinet, GeoTransformer, overlapPredator}, where only the coordinates of the matched points are used to estimate the transformation, the RTUME employs information on both the coordinate values and the feature values, resulting in higher accuracy and robustness of the estimates.

\subsection{Loss Functions} \label{loss-sec}
The loss function adopted to train the dense feature extractor, in a supervised fashion, $\cL=\lambda_1\cL_{pw}+\lambda_2\cL_{\UME}+\lambda_3\cL_{reg}$ is composed of three losses aiming at complementary goals: a point-wise contrastive loss to increase point-level features invariance, a UME-contrastive loss for optimizing the features towards conforming with the UME framework assumptions, and an auxiliary registration loss to guide the feature extractor towards optimizing its performance in the task of point cloud registration.
\subsubsection{Point-wise Contrastive Loss.}
Metric learning, and specifically,  contrastive learning, is a widely used approach for training feature extraction models \cite{FCGF,GCL}. Despite the differences between the two input point clouds, the point-wise contrastive loss aims to enforce similarity between the features of matching points (positive group) while reducing the similarity between non-matching points (negative group).
The employed point-wise loss is an adapted version of the Supervised Contrastive Learning loss (SCL) \cite{khosla2020supervised} for optimizing the feature extraction from 3D point clouds. Let $\cM_{pw}$ be the set of positive matches, which includes all point pairs that are \emph{at most} $\epsilon$ meters apart under the ground-truth transformation. Let $\cN_{pw}^{\bsp}$ be the set of negative pairs of $\bsp \in \cP$, \ie, they are \emph{at least} $R$ away from each other under the ground-truth transformation. The Point-Wise Contrastive Loss is then given by:
\begin{equation}\label{eq:pw_loss}
\cL_{pw} = -\sum_{(\bsp,\bsq)\in\cM_{pw}}\log \frac{\exp\left(\bsf_\bsp^T\bsf_\bsq/\tau\right)}{\exp\left(\bsf_\bsp^T\bsf_\bsq/\tau\right) + \sum_{\bsz\in\cN^{\bsp}_{pw}}\exp\left(\bsf_\bsp^T\bsf_\bsz/\tau\right)}
\end{equation}
where $\bsp\in\cP$ is the anchor point, while $\bsq,\bsz\in \cQ$ are its positive match and a negative match, respectively. $\bsf_{\bsp}$ is the feature vector of $\bsp$ and $\tau$ is a temperature parameter.

\subsubsection{UME Contrastive Loss.}
To optimize the features compatibility with the UME framework, we aim at minimizing the distance on the Grassmann Manifold between local UME descriptors of matching neighborhoods, while maximizing the distance between UME descriptors of non-matching neighborhoods. This is achieved through a novel UME Contrastive Loss, which adapts the SCL loss to manifold metric learning by evaluating \eqref{eq:dpf} between UME local descriptors of the selected keypoints. During the training procedure, we sample $\{\bsp_k\}_{k=1}^K$  from $\widetilde{\cP}$ with radius $R$ neighborhood around each point, such that each neighborhood contains at least $N$ points both in the source point cloud and under the ground-truth transformation, in the target point cloud. The points on the target point cloud are  denoted by $\{\bsq_k\}_{k=1}^K$ (note that in general, $\bsq_k$ is not necessarily an actual point that exists in $\widetilde{\cQ}$ although its neighborhood is). For each point we generate a UME local descriptors $\{\vH_{\bsp_k}\}_{k=1}^{K}, \{\vH_{\bsq_k}\}_{k=1}^{K}$. Let $\cM_{\UME}$ be the set of  UME local descriptors of matching pairs under the ground-truth transformation and $\cN_{\UME}^{\bsp}$ the set of all non-matching neighborhoods to that of $\bsp$ w.r.t. ground truth transformation. The UME Contrastive loss is given by:
\begin{equation}\label{eq:ume_loss}
\cL_{\UME} = -\sum_{(\bsp,\bsq)\in\cM_{\UME}}\log \frac{\exp\left(s_{\bsp\bsq}^{\UME}/\tau\right)}{\exp\left(s_{\bsp\bsq}^{\UME}/\tau\right) + \sum_{\bsz\in\cN^{\bsp}_{\UME}}\exp\left(s_{\bsp\bsz}^{\UME}/\tau\right)}
\end{equation}
where $s_{\bsp\bsq}^{\UME}=1-d_{pF}\left(\vH_{\bsp},\vH_{\bsq}\right)$ is the UME Similarity based on the $d_{pF}$ distance.

\subsubsection{Registration Loss.}
For every matched pair of UME local descriptors we estimate the rigid transformation relating them (as detailed in Sec. \ref{sec:hyp_gen}). To guide the training procedure  towards the task of point cloud registration, we employ an auxiliary loss that ensures the hypothesized transformation estimates from correct matches are accurate. By doing so we tune the feature extractor module to a solution that is implicitly UME compatible as well. We employ the Cube Reprojection Error (CRE) \cite{se3_dist} as our registration loss:
\begin{equation}\label{eq:reg_loss}
\cL_{reg} = \frac{1}{K}\sum_{k=1}^{K}\sum_{\bsp\in C_R}\left\|\left(\widehat{\bsR}_k-\bsR^{GT}\right)\bsp + \left(\widehat{\bst}_k-\bst^{GT}\right)\right\|_2
\end{equation}
$C_R$ is a 3D cube centred at the origin with $R$ length sides, $\{(\widehat{\bsR}_k,\widehat{\bst}_k)\}_{k=1}^K$ are the hypnotized and $(\bsR^{GT},\bst^{GT})$  the ground truth transformations, respectively.

\subsection{Hypothesis Selection}
Once a set of hypotheses is obtained, the best hypothesis is selected by using the  \emph{Point Clouds Feature Correlation Hypothesis Testing} (PC-FCHT) \cite{ojsp}, instead of the commonly employed sample consensus criterion:
Define the feature correlation between two point clouds $\cP$ and $\cQ$ as function of an arbitrary rigid transformation $T\in \SE(3)$  by:
\begin{equation}
\label{eq:pc_corr}
   ( \cP*\cQ)(T) = \sum_{\vp\in\cP}\kappa( f(\vp), f(T^{-1}(\cQ)))
\end{equation}
where $f(\vp)$ is the observation coloring function and $\kappa$ is a weighted correlation function of a single point defined by: $ \kappa( f(\vp), f(\cQ)) =
     \sum_{\vq\in\mathcal{Q}}w_\sigma(\|\vp-\vq\|) f(\vp)^T f(\vq),   $
and $w_\sigma: \mathbb{R}^+\to [0,1]$ is a weight function, inversely related to the distance between given points. Inspired by classic matched filtering detection,  \eqref{eq:pc_corr} is employed to {\it decide} on the best hypothesis among a set of hypothesized transformations in a {\it global} registration framework.
More specifically, let $\mathcal{D} = \{(\bsR_k,\bst_k)\}_{k=1}^K$ be a set of hypothesis estimates. \eqref{eq:pc_corr} is used to measure the quality of transformation estimates, where the goal is to find the transformation  $\widehat{D}$ that maximizes \eqref{eq:pc_corr}, instead of the commonly used consensus size criterion.

\section{Experimental Results}
\label{sec:results}

We evaluated UMERegRobust on the outdoor registration benchmarks of KITTI \cite{KITTI} and nuScenes \cite{nuscenes2019} (Sec. \ref{sec:outdoor}), as well as on the indoor registration benchmark of 3DMatch \cite{3DMatch} (Sec. \ref{sec:indoor}). Each benchmark was compared against a wide range of SOTA registration baselines. Our findings highlight UMERegRobust’s superior performance in most tested scenarios, particularly in extreme cases involving large rotations in outdoor settings, even under strict registration criteria. Additionally, UMERegRobust demonstrated comparable results on the indoor benchmark. Implementation details are introduced in the Supplementary.

\subsection{Outdoor Registration Benchmarks: KITTI \& nuScenes}
\label{sec:outdoor}
\subsubsection{Datasets.} 
We follow the KITTI and nuScenes registration benchmarks as suggested in \cite{GCL}, which includes challenging examples of LiDAR scans that are at most $50m$ apart. Upon analyzing these benchmarks, we observed that most registration problems involve small rotations ($\leq 30^\circ$). Therefore, we propose RotKITTI and RotnuScenes, registration benchmarks specifically designed for scenarios with large real rotations (problems distributed uniformly within $[30^\circ, 180^\circ]$). As shown in Fig. \ref{fig:teaser}, scenarios with large rotations often exhibit significant differences between point clouds due to major differences in their field of view, causing partial overlap and sampling variations between the observations. These types of scenarios are suitable for examining registration methods aimed for  loop closure in SLAM. RotKITTI and RotnuScenes are used solely for testing. Additionally, we report results on the LoKITTI and LonuScenes registration benchmarks as defined in \cite{GCL}. These benchmarks focus on low overlapped scans ($< 30 \%$ overlap) and emphasize large translations with rare occurrences of large rotations (contrary to RotKITTI and RotnuScenes, which include both large rotations and low overlap). Additional details on RotKITTI and RotnuScenes are in the Supplementary.

\subsubsection{Metrics.}
We follow the standard evaluation metrics as defined in \cite{FCGF}. These are Relative Rotation Error (RRE) and Relative Translation Error (RTE). Our main evaluation metric is the Registration Recall (RR) that gather both rotation and translation estimation performance and defined by:
$\RR@(\theta,d) = \frac{1}{N}\sum_{n=1}^{N} [\RRE_n \leq \theta \wedge \RTE_n\leq d]$
where $[\cdot]$ is the Iverson bracket. We define two working points for the RR: \emph{Normal precision}, $\RR@(1.5^\circ, 0.6m)$ and \emph{Strict precision}, $\RR@(1^\circ, 0.1m)$. Evaluating registration performance with high precision is crucial in various tasks, as explained in Sec. \ref{sec:intro}. (Additional details regarding metrics are provided in the Supplementary).

\subsubsection{Results.}
\begin{table}[b]
\caption{Outdoor Registration Benchmarks - Registration Recall $[\%]$}
\label{tab:outdoor}
\centering
\begin{tabular}{lcccccc||cccccc}
\toprule
         & \multicolumn{6}{c }{KITTI Benchmarks} & \multicolumn{6}{c }{nuScenes Benchmarks} \\ 
\cmidrule(rl){2-7} \cmidrule(rl){8-13}
  & \multicolumn{2}{c}{KITTI} & \multicolumn{2}{c}{RotKITTI} & \multicolumn{2}{c}{LoKITTI} & \multicolumn{2}{c}{nuScenes} & \multicolumn{2}{c}{RotnuScenes} & \multicolumn{2}{c}{LonuScenes} \\
  \cmidrule(rl){2-3} \cmidrule(rl){4-5} \cmidrule(rl){6-7} \cmidrule(rl){8-9} \cmidrule(rl){10-11} \cmidrule(rl){12-13}
Method & \quad $N.$ \quad & \quad $S.$ \quad & \quad$ N.$ \quad & \quad $S.$ \quad &\quad$ N.$ \quad & \quad $S.$ \quad &\quad $N.$ \quad & \quad $S.$ \quad &\quad $N.$ \quad & \quad $S.$ \quad &\quad $N.$ \quad & \quad $S.$ \quad  \\ \hline 
FCGF \cite{FCGF}  &75.1  & 73.1  & 11.6  & 3.6  & 17.2  & 6.9  & 58.2  & 37.8  & 5.5 & 5.2  & 1.9  & 0.0   \\ 
Predetor \cite{overlapPredator}  & 88.2  & 58.7  & 41.6  & 35.0  & 33.7  & \underline{28.4}  & 53.9  & 48.1  & 16.5 & 15.7  & 35.6  & 4.2   \\ 
CoFiNet \cite{cofinet}  & 83.2  & 56.4  & 62.5  & 30.1  & 11.2  & 1.0  & 62.3  & 56.1  & 27.0 & \underline{23.6}  & 30.3  & \underline{23.5} \\ 
GeoTrans \cite{GeoTransformer} & 66.3  & 62.6  & \underline{78.5}  & \underline{50.1}  & 37.8  & 7.2  & 70.7  & 37.9  & \underline{34.3} & 13.1  & 48.1  & 17.3   \\ 
GCL \cite{GCL}          & \underline{93.9}  & \underline{78.6}  & 40.1  & 28.8  & \bf{72.3}  & 26.9  & \underline{82.0}  & \underline{67.5}  & 21.0 & 19.6  & \underline{62.3}  & 5.6 \\ \hline
Ours   & \bf{94.3}  & \bf{87.8}  & \bf{81.1}  & \bf{73.3}  & \underline{59.3}  & \bf{30.2}  & \bf{85.5}  & \bf{76.0}  & \bf{51.9} & \bf{39.7}  & \bf{70.8}  & \bf{56.3}  \\ 
\bottomrule
\end{tabular}
\end{table}
Tab. \ref{tab:outdoor} compares various baseline methods on the KITTI and nuScenes registration benchmarks. We report the registration Recall (RR) under Normal precision ($N.$) and Strict precision ($S.$), with the best results in each column highlighted in \textbf{bold} and the second-best results \underline{underlined}. On the regular KITTI benchmark, our method achieves slightly better results under normal precision compared to the previous SOTA. However, under strict precision criteria, we observe a performance gain of $+9\%$.

\begin{figure}[t]
        \begin{minipage}[m]{0.3\linewidth}
        \centering\includegraphics[width=0.99\linewidth]{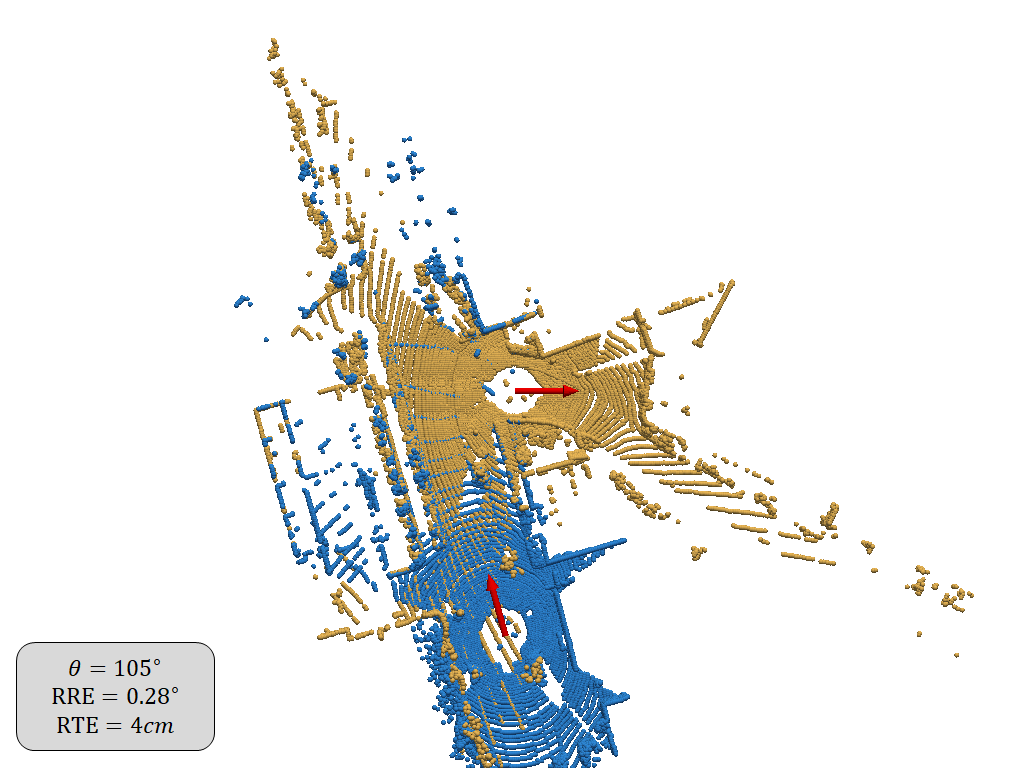}
        \end{minipage}
        \begin{minipage}[m]{0.3\linewidth}
        \centering\includegraphics[width=0.99\linewidth]{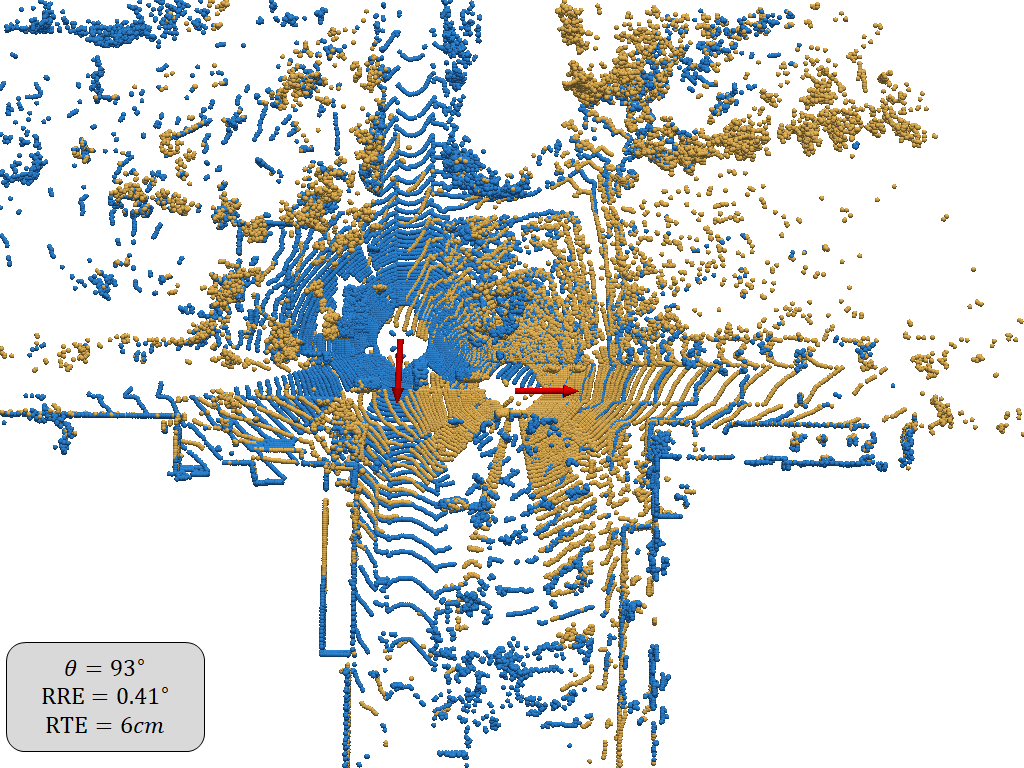}
        \end{minipage}
        \begin{minipage}[m]{0.3\linewidth}
        \centering\includegraphics[width=0.99\linewidth]{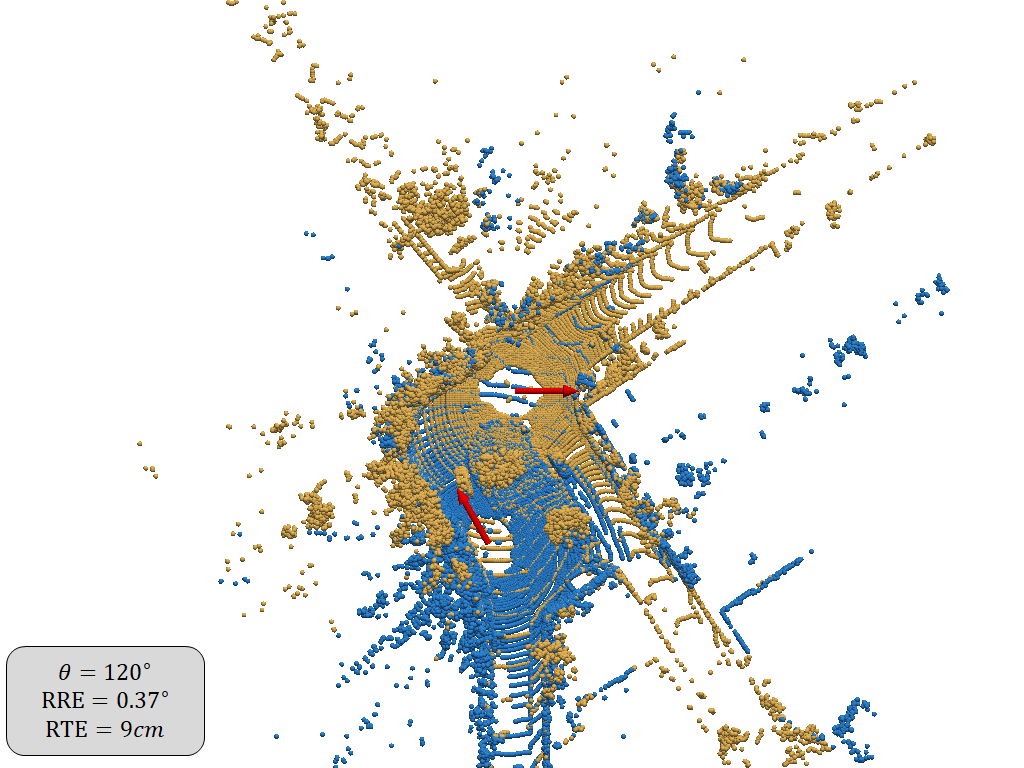}
        \end{minipage}
\caption{Registration results using UMERegRobust. The red arrows depict the vehicle direction at the time the LiDAR scans were acquired. The relative rotation between point clouds ($\theta$), and registration rotation and translation errors (RRE and RTE) are given for each of the examples. Best viewed zoomed in.}\label{fig:qualitative_exmp}
\end{figure}

In the case of RotKITTI, most of the compared methods show significant deterioration, indicating a lack of robustness to rotations. In contrast, UMERegRobust maintains high and stable performance across both the RotKITTI and regular KITTI benchmarks. While GeoTransformer \cite{GeoTransformer} achieves comparable results to our method under normal precision, UMERegRobust outperforms GeoTransformer under strict precision with a notable gain of $+23\%$. Additionally, we observe a significant gain of $+45\%$ over the   GCL \cite{GCL}   SOTA method. Fig. \ref{fig:qualitative_exmp} showcases qualitative examples from the RotKITTI benchmark, highlighting challenging scenarios of large rotations along with our registration results.

Regarding the LoKITTI benchmark, our method achieves the second-best results under normal precision, while GCL maintains the best results. GCL's optimization for low overlap and large translations gives it an inherent advantage on the LoKITTI benchmark compared to other methods. However, our method performs better on low overlap and large rotations and still achieves comparable results on low overlap and large translations, especially under strict precision.

For the nuScenes registration benchmarks, we observe an overall performance degradation in RR values compared to KITTI. This can be attributed to the low-density LiDAR used in the nuScenes dataset (32 beams) \cite{nuscenes2019} compared to the high-density LiDAR (64 beams) \cite{KITTI} in the KITTI dataset. Similar conclusions to these observed on KITTI can be drawn from Tab. \ref{tab:outdoor} regarding the nuScenes and RotnuScenes benchmarks. However, for LonuScenes, UMERegRobust achieves better results compared to GCL, particularly under strict precision. This can be explained by the influence of the SEM, which has a more significant impact on the low-density nuScenes dataset, particularly affecting the LonuScenes examples.

\subsection{Indoor Registration Benchmark: 3DMatch}
\label{sec:indoor}
\begin{figure}[b]
  \begin{minipage}[l]{0.45\linewidth}
  \captionof{table}{3DMatch Benchmark}
\label{tab:indoor}
    \centering
   \begin{tabular}{lcc}
\toprule
Method        & IR [\%]       & RR [\%]   \\
\hline \hline
Lepard \cite{lepard2021} & 55.5 & \underline{93.5} \\
PEAL \cite{peal2023}  & 72.4 & \bf{94.6} \\
YOHO \cite{yoho}  &64.4 & 90.8 \\
RoReg \cite{roreg} & \bf{86.0} & 93.2 \\
E2E \cite{e2e_iros2023} &53.0 & 91.2 \\
FCGF \cite{FCGF} & 56.8 & 85.1        \\ 
Predator \cite{overlapPredator} & 58.0 & 89.0      \\ 
CoFiNet \cite{cofinet}   & 49.8 & 89.3      \\ 
GeoTrans \cite{GeoTransformer} & 71.9 & 92.0     \\ 
\hline
Ours & \underline{79.7} & 93.4      \\ 
\bottomrule
\end{tabular}
  \end{minipage}
  \begin{minipage}[r]{0.45\linewidth}
  \captionof{table}{Ablation Studies}
\label{tab:ablation_study}
    \centering
\centering
\begin{tabular}{p{0.75cm}p{1.5cm}p{1cm}p{1cm}p{1cm}p{1cm}}
\toprule
\multirow{3}{*}{\centering SEM}         & \multirow{3}{=}{\centering UME - Coloring}       & \multicolumn{4}{c}{Registration Recall $[\%]$} \\
\cmidrule(rl){3-6}
& & \multicolumn{2}{c}{KITTI}    & \multicolumn{2}{c}{RotKITTI}  \\
\cmidrule(rl){3-4} \cmidrule(rl){5-6}
& & \centering N. & \centering S. &  \centering N. & \centering S. \tabularnewline
\hline \hline
\centering \xmark & \centering \xmark &   \centering 79.8 &  \centering 67.3 &    \centering 17.5 &   \centering 13.1 \tabularnewline
\centering \cmark & \centering \xmark &   \centering 81.9 &  \centering 68.1 &   \centering 17.5 &   \centering 12.5 \tabularnewline
\centering \xmark & \centering \cmark &   \centering 90.3 &  \centering 74.9 &     \centering 74.4 &   \centering 66.2 \tabularnewline
\centering \cmark & \centering \cmark &   \centering 94.3 &  \centering 87.7 &     \centering 81.0 &   \centering 73.2 \tabularnewline
\hline
\centering \xmark & \centering $-$ &   \centering 93.9 &  \centering 78.6 &   \centering 40.1 &   \centering 28.8 \tabularnewline
\centering \cmark & \centering $-$ &   \centering 84.4 &  \centering 81.4 &   \centering 49.6 &   \centering 48.4 \tabularnewline
\bottomrule
\end{tabular}

  \end{minipage}
\end{figure}
\subsubsection{Dataset.}
We adhere to the 3DMatch registration benchmark as used in prior studies \cite{FCGF, overlapPredator, GeoTransformer}. The dataset comprises 48 scenes for training, 8 scenes for validation, and 8 scenes for testing.

\subsubsection{Metrics.}
The common metrics for the 3DMatch benchmark are Feature Matching Recall (FMR), Inlier Ratio (IR), and Registration Recall (RR). FMR is not reported in this work because it aims to predict method performance under RANSAC, while our method is RANSAC-free. Therefore, we report both IR and RR, with the later as our primary metric. Additional information regarding these metrics can be found in the supplementary material.

\subsubsection{Results.}
Tab. \ref{tab:indoor} presents a comparison between UMERegRobust and various baseline methods, including additional SOTA methods primarily evaluated on indoor scenarios \cite{lepard2021, peal2023, yoho, roreg, e2e_iros2023}. The 3DMatch benchmark shows saturated results for recent methods, with our method achieving comparable results in both the RR and IR metrics.

\subsection{Ablation Studies}
Tab. \ref{tab:ablation_study} presents ablation studies assessing the impact of various components within the UME-compatible coloring module. Specifically, we examine the influence of the \emph{Sampling Equalizer Module} (SEM), the training of the feature extractor using UME-compatible losses, and their combined effect. Models trained with non-UME-compatible coloring use a contrastive loss, as described in \cite{FCGF}. Their performance is compared to the performance obtained using the losses outlined in Sec. \ref{loss-sec}. These properties are evaluated across two categories: the regular KITTI benchmark and the RotKITTI benchmark, with registration recall reported for both Normal precision ($N.$) and Strict precision ($S.$).

When the coloring is not enforced to be UME-compatible, a performance degradation is observed across all categories and precision levels. However, when the SEM is employed, a slight performance improvement is evident. Optimal results are achieved when both UME-compatible coloring and the SEM are utilized, highlighting the synergistic benefits of their combination for the UME framework. An additional ablation study is depicted in the last two lines of Tab. \ref{tab:ablation_study}, where one of the baseline methods (in this case, GCL \cite{GCL}) is used along with the SEM. The SEM did not provide a performance gain for Normal precision but did show some improvement on RotKITTI. Thus, the SEM's contribution to performance is not stand-alone; rather, its integration with the UME maximizes their joint contribution.

%
%
\section{Conclusions}
In this paper, we adopt the Universal Manifold Embedding framework  for the estimation of rigid transformations and extend it, so that it can accommodate scenarios involving partial overlap and differently sampled point clouds.
We extend the UME framework by introducing a
UME-compatible feature extraction method augmented with a unique UME contrastive loss and a sampling equalizer. These components are integrated   into a comprehensive and robust registration pipeline,  named {UMERegRobust}.  UMERegRobust achieves better than state-of-the-art performance on the KITTI benchmark, especially when strict precision  of $(1^\circ, 10cm)$ is considered, and notably outperform SOTA methods on the RotKITTI benchmark where scenarios involving large rotations are considered. Although UMERegRobust outperforms SOTA methods, the  performance measures of UMERegRobust, while closing the gap, still require further improvement in order to handle the considered difficult scenarios, with respect to the strict performance measures required for actual deployment on autonomous robots.

\section*{Acknowledgments}
This work was supported by Israel Innovation Authority under Grant 77887.
\bibliographystyle{splncs04}
\bibliography{ref}

\newpage
\appendix
{\maketitlesupplementary}
\section{Sampling Equalizer Module}
\begin{figure}[ht]
        \begin{minipage}[m]{0.45\linewidth}
        \centering\includegraphics[width=0.99\linewidth]{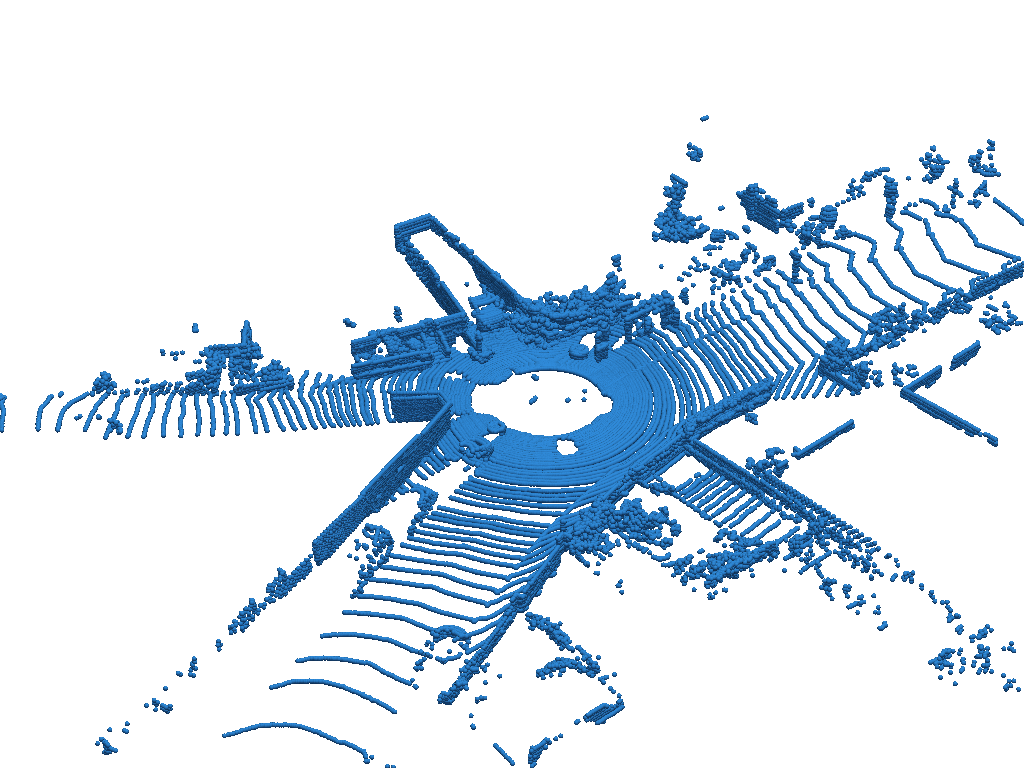}
        \end{minipage}
        \begin{minipage}[m]{0.45\linewidth}
        \centering\includegraphics[width=0.99\linewidth]{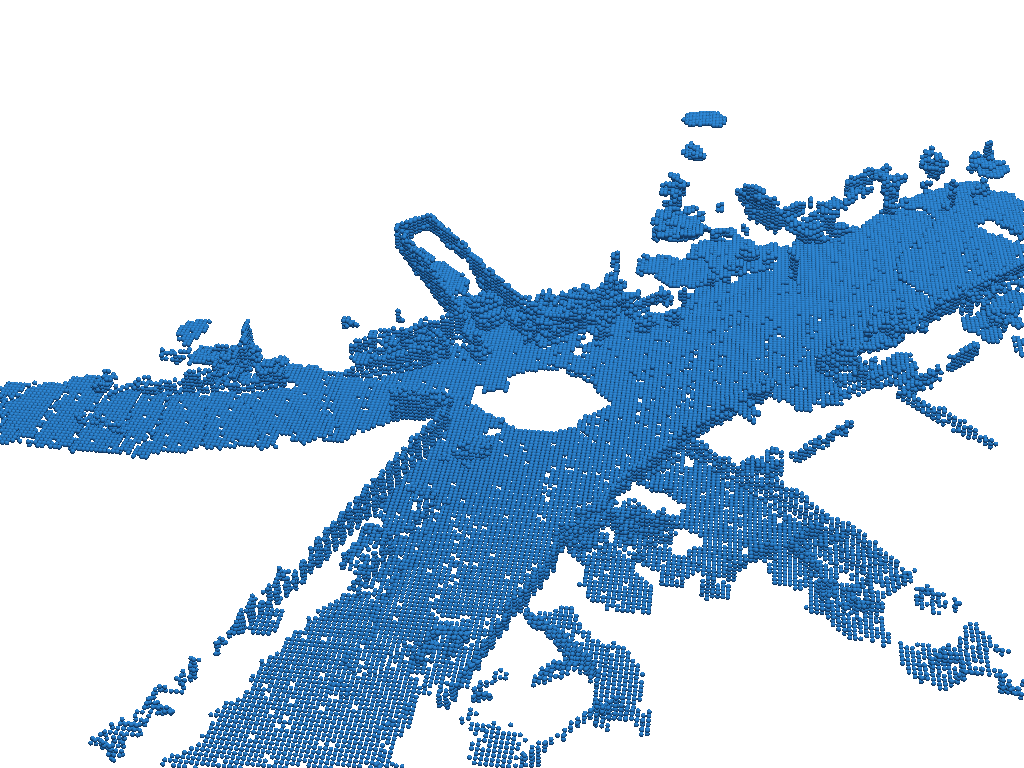}
        \end{minipage}
\caption{Sampling Equalizer Module (SEM) Effect - \textbf{Left:} Raw LiDAR point cloud. \textbf{Right:} Point cloud after SEM.}\label{fig:sem_example}
\end{figure}

The effect of the Sampling Equalizer Module (SEM) is illustrated in Fig. \ref{fig:sem_example}. The final point cloud exhibits considerably more uniform sampling, indicating the reduction of the significant sampling variations typically introduced by the LiDAR sampling pattern, especially at greater distances. Our ablation studies (refer to Sec. 4.3 in the main paper) show that incorporating the SEM with UME-compatible coloring yields the best results.

\section{RotKitti Dataset Preparation}
The RotKITTI is a new registration benchmark created from the original KITTI test sequences. We followed previous works\cite{D3Feat, FCGF, overlapPredator, GeoTransformer, GCL} dataset split, that is 0-5 for training, 6-7 validation and 8-10 for testing. To emphasize the large and challenging characteristics of the rotations, as detailed in Sec. 4.1 of the main paper, we present the relative rotation for each potential problem in the test set using a single-angle representation, based on the ground-truth transformation as evaluated by \eqref{eq:rot_kitti_gen}. Subsequently, we categorized the registration problems into five groups based on their relative rotations: $[30^\circ,60^\circ]$, $[60^\circ,90^\circ]$, $[90^\circ,120^\circ]$, $[120^\circ,150^\circ]$, and $[150^\circ,180^\circ]$. We selected 200 problems from the first category (with the smallest rotations) and 100 problems from each subsequent category. It is important to emphasize that RotKITTI is exclusively used for testing, and no adjustments were made to the training set. Similarly, we created the RotnuScenes registration benchmark from the nuScenes test sequences (as defined in \cite{GCL} train-val-test split).
Additional details are available in our GitHub\footnote{\href{https://github.com/yuvalH9/UMERegRobust}{https://github.com/yuvalH9/UMERegRobust}.} for reproducibility. 
\begin{equation}\label{eq:rot_kitti_gen}
    \RRE=\arccos( 0.5(\Tr(\bsR^{GT})-1))
\end{equation}

\section{Implementation Details}
\subsubsection{Network Architecture.}
\begin{figure}[ht]
        \centering\includegraphics[width=0.99\linewidth]{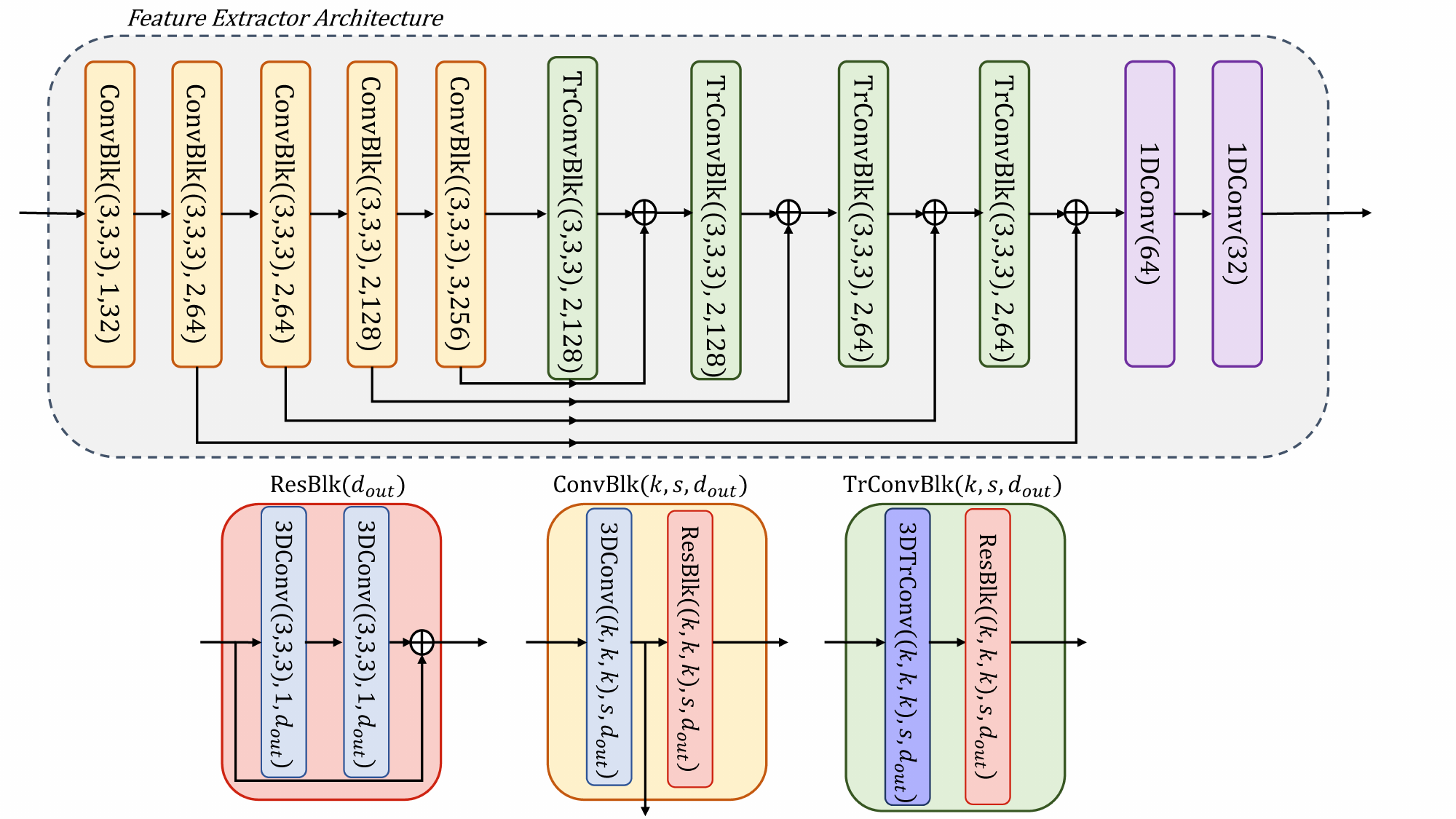}
\caption{Feature Extractor Architecture -  3D sparse convolutions are used with kernel size $k$, stride $s$ and output dimension $d_{out}$. Each convolution layer is followed by Batch Normalization and ReLu, $\oplus$ is a concatenation operation.}\label{fig:dnn_arc}
\end{figure}

We adopt  for the Dense Feature Extractor a similar architecture to that described in \cite{FCGF}, employing sparse 3D convolutions. The feature extractor follows a UNet-based deep neural network (DNN) structure with skip connections. Each convolutional module is paired with Batch Normalization and ReLU activation (except for the last layer), and the output features are normalized to unit norm. We consider the UME local descriptor radius when designing the DNN receptive field resulting in different kernel sizes and stride values compared to \cite{FCGF}. The complete architecture is illustrated in Fig. \ref{fig:dnn_arc}.

\subsubsection{UME Descriptors.}
For the generation of the Local UME descriptors on the outdoor data, we utilized neighborhoods with a radius of $R=5 m$, ensuring they include at least 300 points, and for the indoor data we utilize neighborhoods with $R=0.3m$ and at least 100 points. We selected the set $\{w_m(\bsx)\}_{m=1}^{M}$ functions to be $\{\langle \bse_m,\bsx\rangle\}_{m=1}^{M}$ which extracts the $m$-th entry of $\bsx$, with $M=32$ ($M$ is also the output dimension of the Coloring module). Regarding the SEM details, we used a frozen NKSR \cite{NKSR}, with its official weights, selecting a grid size to be $\rho=30cm$ for outdoor scenarios (KITTI and nuScenes) and $\rho=2.5cm$ for indoor scenarios (3DMatch).

\subsubsection{UME $d_{pF}$ Calculation.}
The distance between two local UME descriptors $\vH_{\bsp},\vH_{\bsq}$ is computed using the $d_{pF}$ distance, as described in Equation (3) in the main paper. To calculate this distance, we require the projection matrices $\vP_{\vH_{\bsp}}, \vP_{\vH_{\bsq}}$ onto the column spaces of $\vH_{\bsp},\vH_{\bsq}$, respectively. Utilizing \emph{QR-Decomposition} on each local UME descriptor, we obtain the orthogonal matrix $\vQ$, from which we derive the corresponding projection matrices $\vP_{\vH_{\bsp}}=\vQ_{\vH_{\bsp}}\vQ_{\vH_{\bsp}}^T$ and $\vP_{\vH_{\bsq}}=\vQ_{\vH_{\bsq}}\vQ_{\vH_{\bsq}}^T$ for $\vH_{\bsp},\vH_{\bsq}$, respectively. We note that the QR-Decomposition is differentiable for full-rank matrices, making it suitable for back-propagation.

\subsubsection{Loss and Training Configurations.}
We assigned weighting coefficients $\lambda_1=0.5$, $\lambda_2=0.5$, and $\lambda_3=0.25$ to the point-wise loss, UME loss, and registration loss, respectively. The temperature parameter $\tau$ was set to $\tau=0.1$ for both the point-wise loss ($\cL_{pw}$) and UME loss ($\cL_{\UME}$), yielding the best results. Negative pairs for the point-wise loss were selected as points separated by at least $5m$. For the UME loss, point cloud down-sampled versions $\widehat{\cP}$ and $\widehat{\cQ}$ contained $K=256$ points, representing the number of keypoints used for training. The registration loss ($\cL_{reg}$) employed a cube with side length of $30m$, approximately equivalent to the radius of the dense area of the LiDAR measurements. We trained our model using the ADAM optimizer for 100 epochs with a learning rate of $1e^{-4}$ and a batch size of 8. Augmentations were applied only for rotations during training.

\subsubsection{Baselines Comparison.}
We follow the baseline evaluations as presented in the original papers. Some methods were evaluated on indoor-only data \cite{lepard2021, peal2023, yoho, e2e_iros2023}, while \cite{GCL} was evaluated on outdoor-only data. For the outdoor benchmark, we adopt the train-test split as in \cite{GCL}; therefore, we retrained \cite{FCGF, overlapPredator, cofinet, GeoTransformer} on the relevant KITTI and nuScenes train sets. Since some of the compared methods include a refinement step using ICP \cite{ICP}, all results are reported after applying ICP. When performed without ICP, all methods demonstrate performance degradation, while ours shows only a slight degradation of $2\%$ to $4\%$. For evaluation, we used $K=5000$ matches (for our method it is equivalent to the number of tested hypotheses contrary to RANSAC based methods) similar to other methods.

\subsubsection{Time Analysis.}
All training and evaluations are done on NVIDIA RTX A6000 (48GB) GPU and 12th Gen Intel(R) Core i7-12700K CPU.
Following \cite{GCL, GeoTransformer} time analysis methodology: UME coloring module takes 0.02s; UME hypotheses generation requires 0.37s; and  hypotheses selection consumes 0.6s; yielding total inference time of $\approx 1$ sec. Measurements were conducted for 5K matches on point cloud sizes of 30K points from the KITTI test set. Our method demonstrates comparable or even superior time performance compared to RANSAC-based methods. Moreover,  code optimization can significantly reduce computation times.

\section{Evaluation Metrics}
For the outdoor registration benchmarks we followed the standart  metrics of Relative Rotation Error (RRE), Relative Translation Error (RTE) and Registration Recall (RR) as given in (\ref{eq:rre}-\ref{eq:rr}).
\begin{align}
&\RRE=\arccos\left(0.5\left({\Tr\left(\widehat{\bsR}^T\bsR^{GT}\right)-1}\right)\right)  \label{eq:rre}\\
    &\RTE=\| \widehat{\bst}-\bst^{GT} \|_2  \label{eq:rte}\\
    &\RR@(\theta,d) = \frac{1}{N}\sum_{n=1}^{N} \left[\RRE_n \leq \theta \wedge \RTE_n\leq d\right]  \label{eq:rr}
\end{align}
where $(\bsR^{GT}, \bst^{GT})$ are the GT rotation and translation, $(\widehat{\bsR}, \widehat{\bst})$ are the estimated rotation and translation and $(\theta,d)$ are the $\RRE$ and $\RTE$ angle and translation errors respectively which the $\RR$ is evaluated at.

For the indoor registration benchmark, we followed the common metrics of Inlier Ratio (IR) and registration Recall (RR), defined by:
\begin{align}
&\IR = \frac{1}{|\cC_{pred}|}\sum_{\bsp, \bsq \in \cC_{pred}} \left[\left\|\bsR^{GT}\bsp + \bst^{GT} -\bsq \right\|_2 < \tau_1 \right]  \label{eq:ir} \\
&E^{RMSE}_i = \sqrt{\frac{1}{|\cC^{GT}_{i}|}\sum_{\bsp^*, \bsq^*\in\cC^{GT}_i}\left\|\widehat{\bsR}_i\bsp^* + \widehat{\bst}_i -\bsq^* \right\|_2} \\
&\RR = \frac{1}{N}\sum_{i=1}^{N}\left[E^{RMSE}_i < \tau_2 \right]
\end{align}
where $\cC_{pred}$ is the set of estimated correspondences, $\cC^{GT}_i$ is the set of correct  correspondences (defined using the GT transformation) for the $i$ registration problem, $(\widehat{\bsR}, \widehat{\bst})$ are the estimated rotation and translation and $[\cdot]$ is the Iverson bracket. $\tau_1=10cm, \tau_2=20cm$  according to the 3DMatch benchmark.

\section{Additional Results}
\subsubsection{KITTI.}
Tab. \ref{tab:kitti_range} shows the registration performance across different categories of the KITTI dataset, where $[b1, b2]$ denotes the distance in meters between the measured point clouds. As the distance between measurements increases, the overlap between observations decreases, creating a more challenging registration problem. At low distances of $[5,20]$, most methods reach saturation at normal precision. In the mid-range distance of $[20,30]$, UMERegRobust achieves slightly better performance compared to GCL and GeoTransformer at normal precision, with more significant gains relative to other methods. Notably, there is a significant difference compared to all methods at strict precision, with UMERegRobust exhibiting a gain of $+26\%$ over GCL. Conversely, at far measurements $[30,50]$, GCL performs better than our method at normal precision. However, when tested at strict precision, our method demonstrates significant gain in the $[30,40]$ range and slightly better performance in the $[30,50]$ range. Furthermore, the performance drop caused by tightening the precision criteria of our method is $23\%$, compared to $40\%$ for GCL.
\begin{table}[]
\caption{KITTI Benchmark - Registration Recall $[\%]$}
\label{tab:kitti_range}
\centering
\begin{tabular}{lcccccccccccc}
\toprule
Method         & \multicolumn{2}{c }{All}                                  & \multicolumn{2}{c }{[5,10]}                               & \multicolumn{2}{c }{[10,20]}                              & \multicolumn{2}{c }{[20,30]}                              & \multicolumn{2}{c }{[30,40]}                              & \multicolumn{2}{c }{[40,50]}         \\ \cmidrule(rl){2-3} \cmidrule(rl){4-5} \cmidrule(rl){6-7} \cmidrule(rl){8-9} \cmidrule(rl){10-11} \cmidrule(rl){12-13}
               & \multicolumn{1}{c }{\quad N. \quad} & \multicolumn{1}{c }{\quad S. \quad} & \multicolumn{1}{c }{\quad N.\quad} & \multicolumn{1}{c }{\quad S. \quad} & \multicolumn{1}{c }{\quad N. \quad} & \multicolumn{1}{c }{\quad S. \quad} & \multicolumn{1}{c }{\quad N. \quad} & \multicolumn{1}{c }{\quad S. \quad} & \multicolumn{1}{c }{\quad N. \quad} & \multicolumn{1}{c }{\quad S. \quad} & \multicolumn{1}{c }{\quad N. \quad} & \quad S. \quad \\ \hline \hline
FCGF   \cite{FCGF}        & \multicolumn{1}{c }{75.1}  & \multicolumn{1}{c }{73.1}  & \multicolumn{1}{c }{97.0}  & \multicolumn{1}{c }{94.9}  & \multicolumn{1}{c }{85.4}  & \multicolumn{1}{c }{80.1}  & \multicolumn{1}{c }{54.1}  & \multicolumn{1}{c }{52.8}  & \multicolumn{1}{c }{25.0}  & \multicolumn{1}{c }{32.8}  & \multicolumn{1}{c }{14.3}  & 9.1   \\
Predetor \cite{overlapPredator}      & \multicolumn{1}{c }{88.2}  & \multicolumn{1}{c }{58.7}  & \multicolumn{1}{c }{99.3}  & \multicolumn{1}{c }{73.1}  & \multicolumn{1}{c }{\underline{96.8}}  & \multicolumn{1}{c }{60.5}  & \multicolumn{1}{c }{90.2}  & \multicolumn{1}{c }{48.7}  & \multicolumn{1}{c }{60.6}  & \multicolumn{1}{c }{32.6}  & \multicolumn{1}{c }{26.7}  & 14.4  \\
CoFiNet \cite{cofinet}       & \multicolumn{1}{c }{83.2}  & \multicolumn{1}{c }{56.4}  & \multicolumn{1}{c }{\bf 99.8}  & \multicolumn{1}{c }{86.0}  & \multicolumn{1}{c }{96.1}  & \multicolumn{1}{c }{61.3}  & \multicolumn{1}{c }{73.0}  & \multicolumn{1}{c }{14.0}  & \multicolumn{1}{c }{41.1}  & \multicolumn{1}{c }{3.5}   & \multicolumn{1}{c }{13.5}  & 0.0   \\
GeoTrans \cite{GeoTransformer} & \multicolumn{1}{c }{66.3}  & \multicolumn{1}{c }{62.6}   & \multicolumn{1}{c }{97.9}  & \multicolumn{1}{c }{78.8}  & \multicolumn{1}{c }{88.3}  & \multicolumn{1}{c }{73.3}  & \multicolumn{1}{c }{92.5}  & \multicolumn{1}{c }{49.2}  & \multicolumn{1}{c }{72.3}  & \multicolumn{1}{c }{24.8}  & \multicolumn{1}{c }{38.7}  & 5.4   \\
GCL \cite{GCL}           & \multicolumn{1}{c }{\underline{93.9}}  & \multicolumn{1}{c }{\underline{78.6}}  & \multicolumn{1}{c }{98.4}  & \multicolumn{1}{c }{\underline{96.1}}  & \multicolumn{1}{c }{96.1}  & \multicolumn{1}{c }{\underline{83.0}}  & \multicolumn{1}{c }{\underline{94.1}}  & \multicolumn{1}{c }{\underline{62.4}}  & \multicolumn{1}{c }{\bf 87.6}  & \multicolumn{1}{c }{\underline{49.6}}  & \multicolumn{1}{c }{\bf 67.6}  & \underline{26.1}  \\
\hline
UMERegRobust   & \multicolumn{1}{c }{\bf 94.3}  & \multicolumn{1}{c }{\bf 87.8}  & \multicolumn{1}{c }{\underline{99.7}}  & \multicolumn{1}{c }{\bf 98.5}  & \multicolumn{1}{c }{\bf 99.1}  & \multicolumn{1}{c }{\bf 95.5}  & \multicolumn{1}{c }{\bf 97.5}  & \multicolumn{1}{c }{\bf 88.5}  & \multicolumn{1}{c }{\underline{80.8}}  & \multicolumn{1}{c }{\bf 62.4}  & \multicolumn{1}{c }{\underline{58.5}}  & {\bf 30.3}  \\
\bottomrule
\end{tabular}
\end{table}

\subsubsection{RotKITTI.}
Tab. \ref{tab:Table1-RotKitti} provides  results on the RotKITTI benchmark, categorized into 5 ranges based on the relative rotation angle of the problem denoted as $[\theta_1,\theta_2]$ in degrees, reported for both normal ($N.$) and strict ($S.$) registration precision. In Fig. \ref{fig:qualitative_examples}, qualitative examples from the RotKITTI benchmark are showcased, highlighting the challenging scenarios of large relative rotations alongside our registration results.
Large relative rotations can occur due to various factors, such as the ego-vehicle approaching a revisited location from a different direction (e.g., arriving from the opposite lane or another side of a junction), or when the vehicle makes turns.
From Tab. \ref{tab:Table1-RotKitti}, it is evident that the majority of the compared methods exhibit significant deterioration on the RotKITTI benchmark, indicating their lack of robustness to rotations. In contrast, UMERegRobust maintains high and stable performance across both the RotKITTI and regular KITTI benchmarks.
GeoTransformer, however, achieves comparable results to our method and even leads in the $[90^\circ,120^\circ]$ range with normal precision. However, with strict precision, UMERegRobust outperforms GeoTransformer with a performance gain of $+25\%$. Additionally, our method demonstrates higher precision across all categories with a much smaller performance drop from normal to strict precision compared to others, resulting in a total gain of $+45\%$ over the SOTA GCL method and a total of $+23\%$ over the second best GeoTransformer.

\begin{table}[]
\caption{RotKITTI Benchmark - Registration Recall $[\%]$}
\label{tab:Table1-RotKitti}
\centering
\begin{tabular}{lcccccccccccc}
\toprule
Method         & \multicolumn{2}{c }{All}                                  & \multicolumn{2}{c }{$[30^\circ,60^\circ]$}                              & \multicolumn{2}{c }{$[60^\circ,90^\circ]$}                              & \multicolumn{2}{c }{$[90^\circ,120^\circ]$}                             & \multicolumn{2}{c }{$[120^\circ,150^\circ]$}                            & \multicolumn{2}{c }{$[150^\circ,180^\circ]$}       \\ \cmidrule(rl){2-3} \cmidrule(rl){4-5} \cmidrule(rl){6-7} \cmidrule(rl){8-9} \cmidrule(rl){10-11} \cmidrule(rl){12-13}
               & \multicolumn{1}{c }{\quad N. \quad} & \multicolumn{1}{c}{\quad S. \quad} & \multicolumn{1}{c}{\quad N. \quad} & \multicolumn{1}{c }{\quad S. \quad} & \multicolumn{1}{c }{\quad N. \quad} & \multicolumn{1}{c }{\quad S. \quad} & \multicolumn{1}{c }{\quad N. \quad} & \multicolumn{1}{c }{\quad S. \quad} & \multicolumn{1}{c }{\quad N. \quad} & \multicolumn{1}{c }{\quad S. \quad} & \multicolumn{1}{c }{\quad N. \quad} & \quad S. \quad \\ \hline \hline
FCGF\cite{FCGF}           & \multicolumn{1}{c }{11.6}  & \multicolumn{1}{c }{3.6}   & \multicolumn{1}{c }{36.5}  & \multicolumn{1}{c }{31.0}  & \multicolumn{1}{c }{2.0}   & \multicolumn{1}{c }{2.0}   & \multicolumn{1}{c }{0.0}   & \multicolumn{1}{c }{0.0}   & \multicolumn{1}{c }{0.0}   & \multicolumn{1}{c }{0.0}   & \multicolumn{1}{c }{1.0}   & 1.0   \\
Predetor\cite{overlapPredator}       & \multicolumn{1}{c }{41.6}  & \multicolumn{1}{c }{35.0}     & \multicolumn{1}{c }{73.4}  & \multicolumn{1}{c }{\underline{72.4}}  & \multicolumn{1}{c }{36.6}  & \multicolumn{1}{c }{36.6}  & \multicolumn{1}{c }{20.5}  & \multicolumn{1}{c }{17.6}  & \multicolumn{1}{c }{16.0}  & \multicolumn{1}{c }{15.0}  & \multicolumn{1}{c }{34.0}  & 33.5  \\
CoFiNet\cite{cofinet}       & \multicolumn{1}{c }{62.5}  & \multicolumn{1}{c }{30.1}  & \multicolumn{1}{c }{67.1}  & \multicolumn{1}{c }{37.5}  & \multicolumn{1}{c }{50.5}  & \multicolumn{1}{c }{17.1}  & \multicolumn{1}{c }{47.0}  & \multicolumn{1}{c }{10.7}  & \multicolumn{1}{c }{56.0}  & \multicolumn{1}{c }{34.0}  & \multicolumn{1}{c }{86.0}  & 43.9  \\
GeoTrans\cite{GeoTransformer} & \multicolumn{1}{c }{\underline{78.5}}   & \multicolumn{1}{c }{\underline{50.1}}  & \multicolumn{1}{c }{89.2}  & \multicolumn{1}{c }{66.3}  & \multicolumn{1}{c }{\underline{70.2}}  & \multicolumn{1}{c }{\underline{40.5}}  & \multicolumn{1}{c }{\bf 68.6}  & \multicolumn{1}{c }{\underline{24.5}}  & \multicolumn{1}{c }{\underline{66.0}}  & \multicolumn{1}{c }{\underline{53.0}}  & \multicolumn{1}{c }{\underline{88.0}}  & \underline{51.0}  \\
GCL\cite{GCL}            & \multicolumn{1}{c }{40.1}  & \multicolumn{1}{c }{28.8}  & \multicolumn{1}{c }{\underline{89.4}}  & \multicolumn{1}{c }{62.9}  & \multicolumn{1}{c }{52.9}  & \multicolumn{1}{c }{38.9}  & \multicolumn{1}{c }{9.0}   & \multicolumn{1}{c }{7.9}   & \multicolumn{1}{c }{0.0}   & \multicolumn{1}{c }{0.0}   & \multicolumn{1}{c }{0.0}   & 0.0   \\
\hline
UMERegRobust  & \multicolumn{1}{c }{\bf 81.1}  & \multicolumn{1}{c }{\bf 73.3}  & \multicolumn{1}{c }{\bf 92.2}  & \multicolumn{1}{c }{\bf 83.3}  & \multicolumn{1}{c }{\bf 75.0}  & \multicolumn{1}{c }{\bf 64.0}  & \multicolumn{1}{c }{\underline{62.4}}  & \multicolumn{1}{c }{\bf 49.5}  & \multicolumn{1}{c }{\bf 71.9}  & \multicolumn{1}{c }{\bf 67.7}  & \multicolumn{1}{c }{\bf 92.8}  & {\bf 91.8}  \\
\bottomrule
\end{tabular}
\end{table}

\subsubsection{Mean Rotation and Translation Errors.}

\begin{table}[b]
\caption{Mean Rotation and Translation Errors - KITTI, LoKITTI and RotKITTI}
\label{tab:mean_errs}
\centering
\begin{tabular}{lcccccc}
\toprule
\multirow{2}{*}{Method}         & \multicolumn{2}{c}{KITTI}       & \multicolumn{2}{c}{LoKITTI}         & \multicolumn{2}{c}{RotKITTI} \\ \cmidrule(rl){2-3} \cmidrule(rl){4-5} \cmidrule(rl){6-7}  
               & $m\RRE[^\circ] \downarrow$   &$m\RTE[m] \downarrow$ & $m\RRE[^\circ] \downarrow$   & $m\RTE[m] \downarrow$ & $m\RRE[^\circ] \downarrow$      & $m\RTE[m] \downarrow$      \\ \hline \hline
FCGF  \cite{FCGF}   & 5.67 & 4.33   & 40.24 & 33.46    & 91.08  & 26.09  \\ 
Predator \cite{overlapPredator}   & 3.32 & 2.59   & 32.05 & 24.63    & 62.00  & 19.85  \\ 
CoFiNet \cite{cofinet}   & 4.26 & 3.75   & 30.79 & 29.28    & 26.76  & 11.60  \\ 
GeoTrans \cite{GeoTransformer}   & \underline{3.13} & 2.27   & 33.02 & 21.59    & \underline{11.44}  & \underline{6.49}  \\ 
GCL  \cite{GCL}   & 3.46 &\underline{1.81}    & \underline{27.63} & \underline{16.15}    & 73.70  & 18.57  \\ 
\hline
UMERegRobust & \textbf{2.08} & \textbf{1.53}   & \textbf{10.02} & \textbf{6.06}    & \textbf{10.96}  & \textbf{5.85}  \\ 
\bottomrule
\end{tabular}
\end{table}
In Tab. \ref{tab:mean_errs}, we examine the mean relative rotation and relative translation errors of various methods across each benchmark: KITTI, LoKITTI, and RotKITTI. Unlike other studies where mRRE and mRTE are solely reported on \emph{successfully} registered problems (\ie problems solved below a fixed value of rotation and translation errors), we report mean errors on \emph{all} problems. This allows for a more comprehensive evaluation of a registration method, considering its performance in all cases, including failures.

From Tab. \ref{tab:mean_errs}, it is evident that the proposed method achieves the best  results in all three benchmarks. While the performance of all methods appears similar on the KITTI benchmark, our method demonstrates a slight performance gain. Further analysis of Tab. 1 in the main paper reveals that while few large errors contribute to increased total mean error values, our method exhibits higher precision, indicating that its larger errors are less significant than those of other methods.
In comparison to the KITTI benchmark, both LoKITTI and RotKITTI show significantly higher  mean error values for all methods, while ours consistently achieves the lowest results with a substantial gain.

\subsubsection{Cube Reprojection Error.}
\begin{figure}[t]
\begin{minipage}{0.45\linewidth}
    \centering
    \includegraphics[width=1\linewidth]{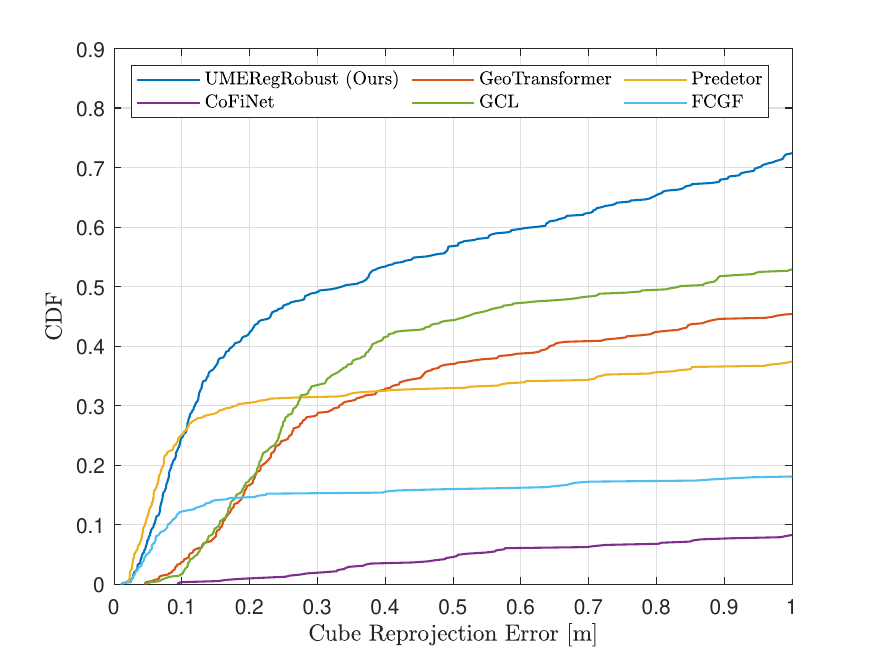}
  \end{minipage}
  \begin{minipage}{0.45\linewidth}
    \centering
    \includegraphics[width=1\linewidth]{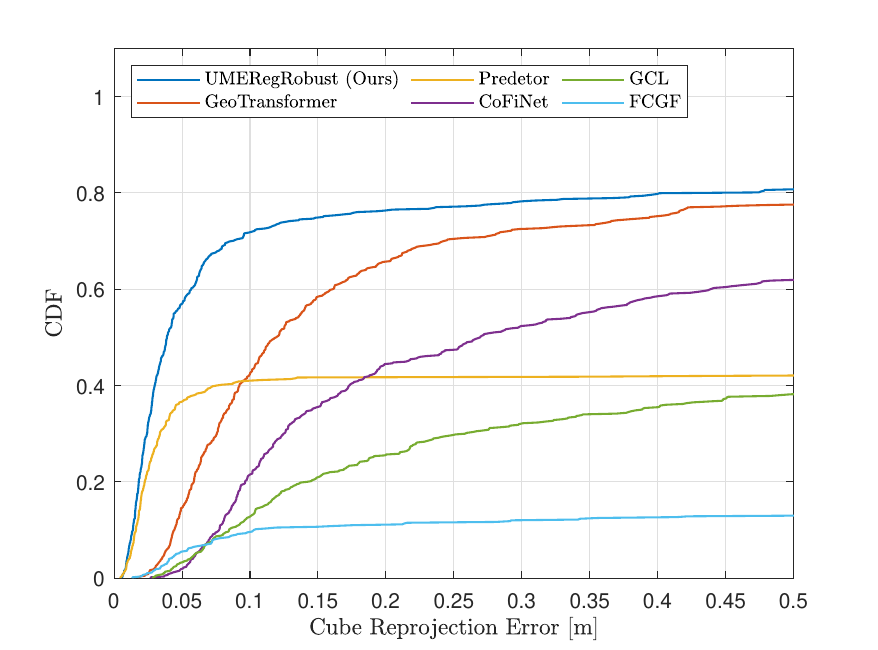}
  \end{minipage}
\caption{CDF of the CRE on LotKITTI (Left). CDF of the CRE on RotKITTI (Right).}
\label{fig:cube_metric_kitti}
\end{figure}

The primary evaluation metric, Registration Recall, assesses joint rotation and translation estimation performance in a strict fashion. A more flexible measure of joint performance can be evaluated using the Cube Reprojection Error (CRE) \cite{se3_dist}. The CRE, as described in (\ref{eq:cre}), quantifies the reprojection error over the vertices of a 3D unit cube between the estimated and ground-truth transformations. This metric integrates both rotation and translation performance into a single scalar value, facilitating further performance analysis, such as examining its Cumulative Distribution Function (CDF).
\begin{equation}\label{eq:cre}
\CRE = \frac{1}{8}\sum_{\bsp\in C_1}\left\|\left(\widehat{\bsR}-\bsR^{GT}\right)\bsp + \left(\widehat{\bst}-\bst^{GT}\right)\right\|_2
\end{equation}
$C_1$ is a 3D unit cube, $(\widehat{\bsR},\widehat{\bst}),(\bsR^{GT}, \bst^{GT})$ are the estimated and the GT transformations, respectively.
We compared the CRE CDF of the baseline methods with ours, on the LoKITTI (Fig. \ref{fig:cube_metric_kitti} Left), and on the RotKITTI (Fig. \ref{fig:cube_metric_kitti} Right) registration benchmarks. From these figures, it is evident that UMERegRobust outperforms others, starting from a very low error threshold of a few centimetres.

\subsection{Qualitative Examples}
\begin{figure}[]
\scriptsize
\hrule
        \centerline{Ground Truth}
        \begin{minipage}[m]{0.22\linewidth}
        \centerline{$\theta=52.3^\circ$}
        \centerline{$t=40.5m$}
        \centering\includegraphics[width=0.99\linewidth]{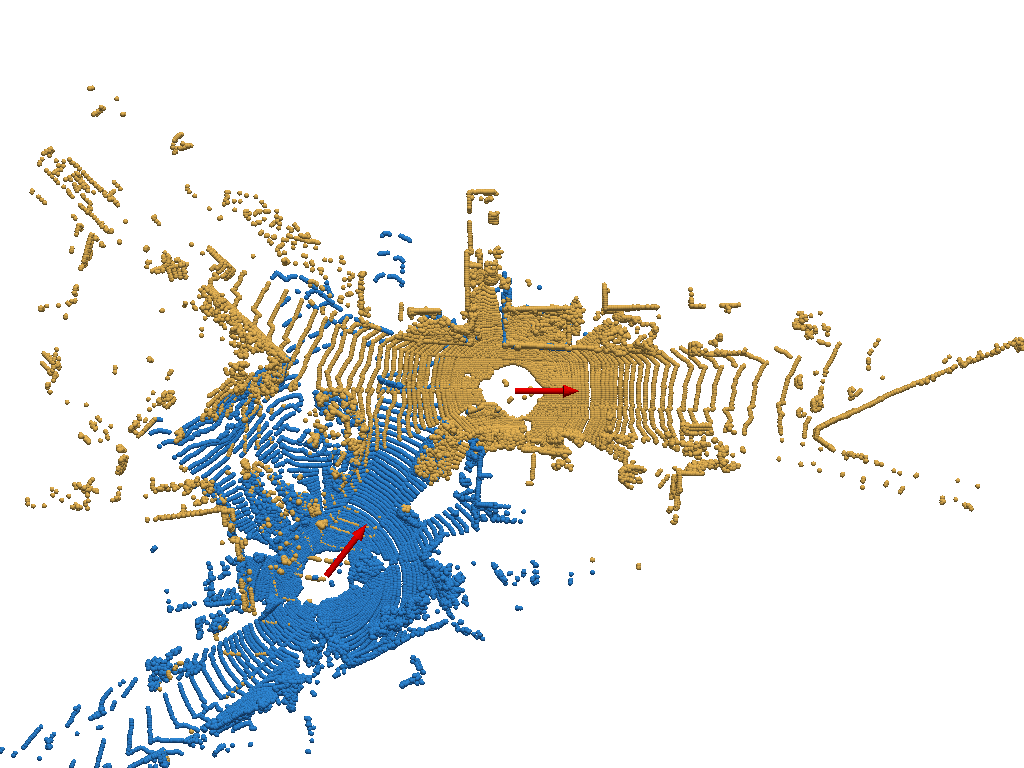}
        \end{minipage}
        \begin{minipage}[m]{0.24\linewidth}
        \centerline{$\theta=54.2^\circ$}
        \centerline{$t=38.2$}
        \centering\includegraphics[width=0.99\linewidth]{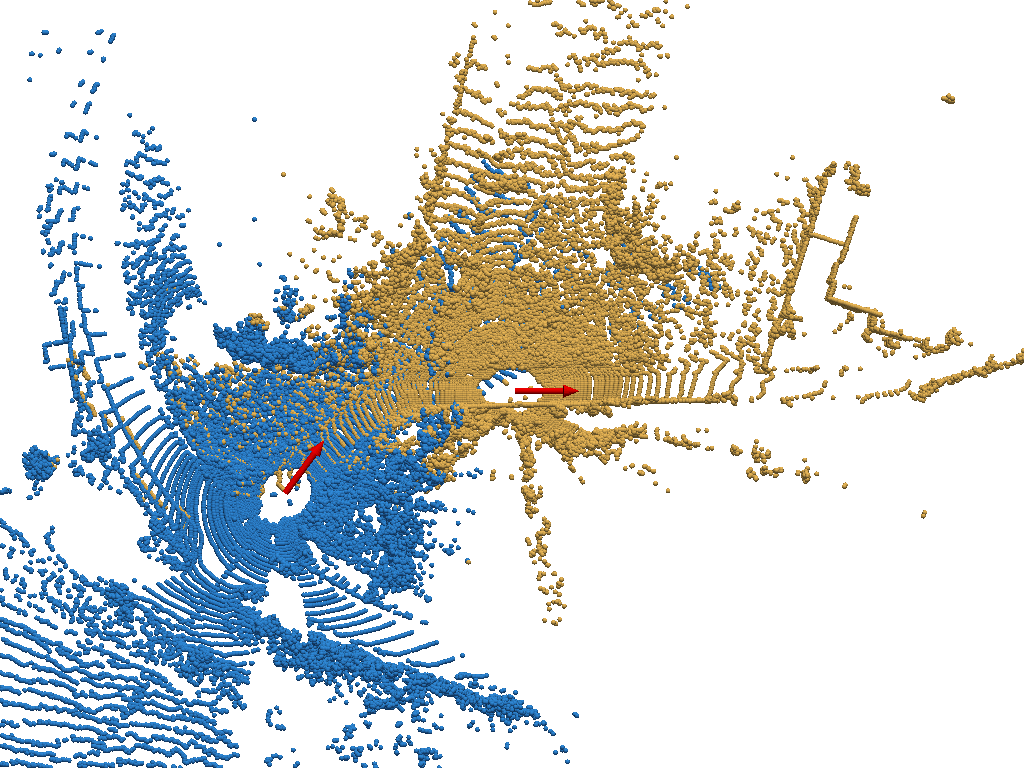}
        \end{minipage}
        \begin{minipage}[m]{0.24\linewidth}
        \centerline{$\theta=102.8^\circ$}
        \centerline{$t=31.0m$}
        \centering\includegraphics[width=0.99\linewidth]{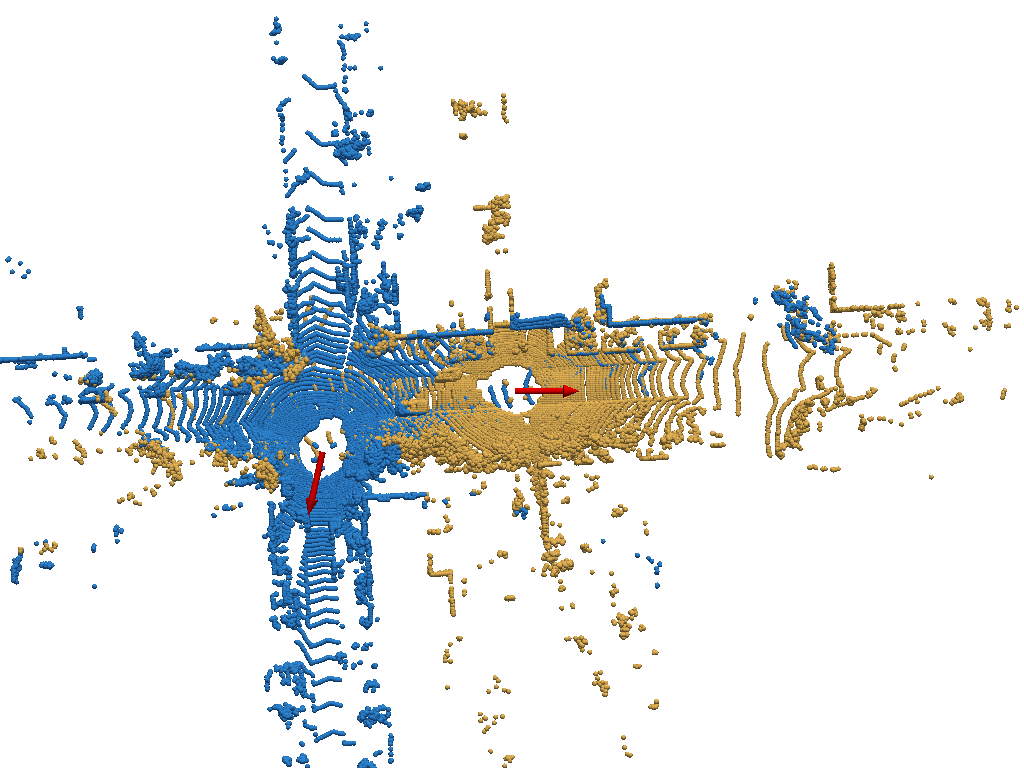}
        \end{minipage}
        \begin{minipage}[m]{0.24\linewidth}
        \centerline{$\theta=97.2^\circ$}
        \centerline{$t=32.8m$}
        \centering\includegraphics[width=0.99\linewidth]{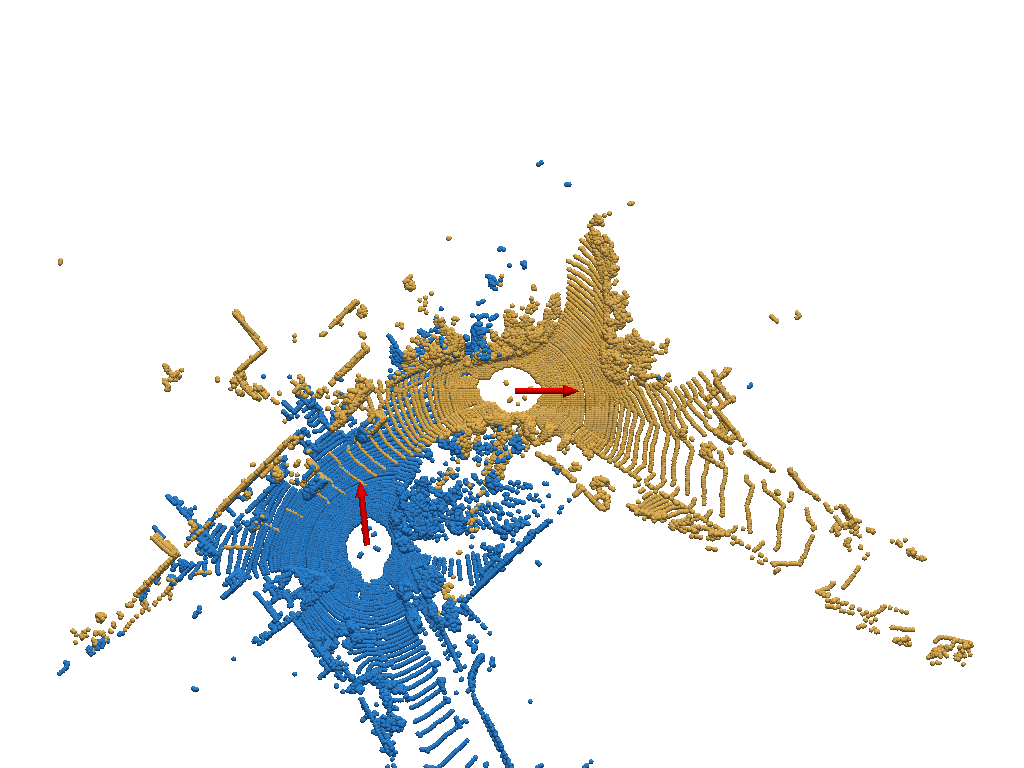}
        \end{minipage}
        \hrule
        \centerline{Ours}
        \begin{minipage}[m]{0.24\linewidth}
        \centerline{$\RRE=0.2^\circ$}
        \centerline{$\RTE=0.01m$}
        \centering\includegraphics[width=0.99\linewidth]{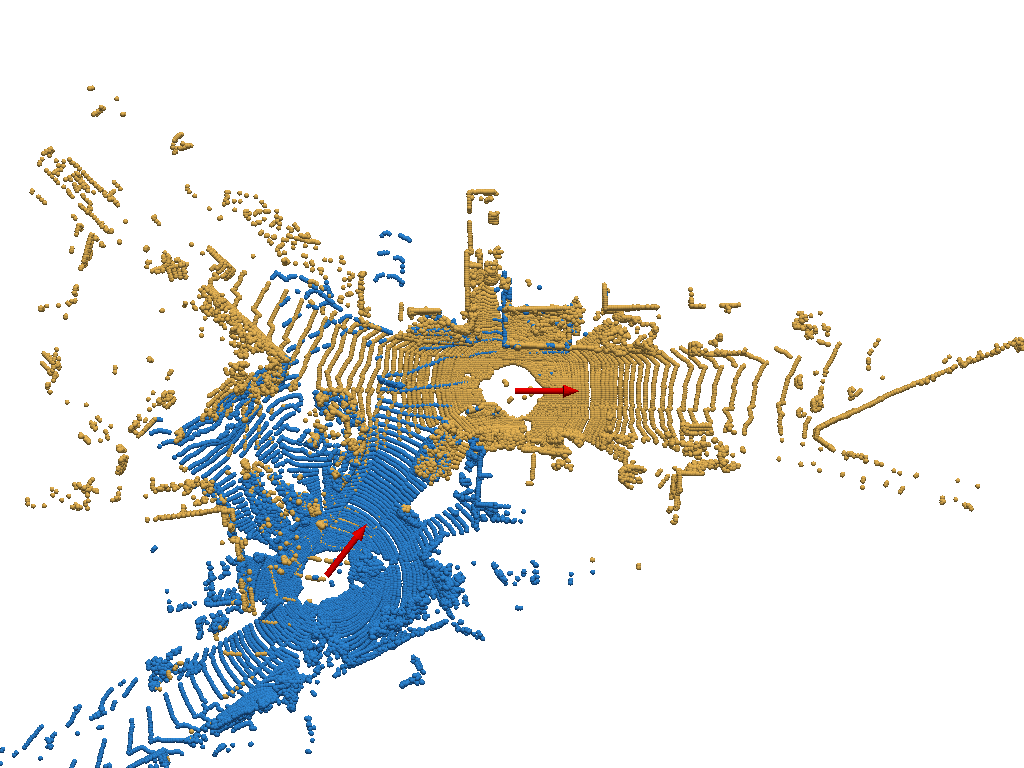}
        \end{minipage}
        \begin{minipage}[m]{0.24\linewidth}
        \centerline{$\RRE=0.4^\circ$}
        \centerline{$\RTE=0.48m$}
        \centering\includegraphics[width=0.99\linewidth]{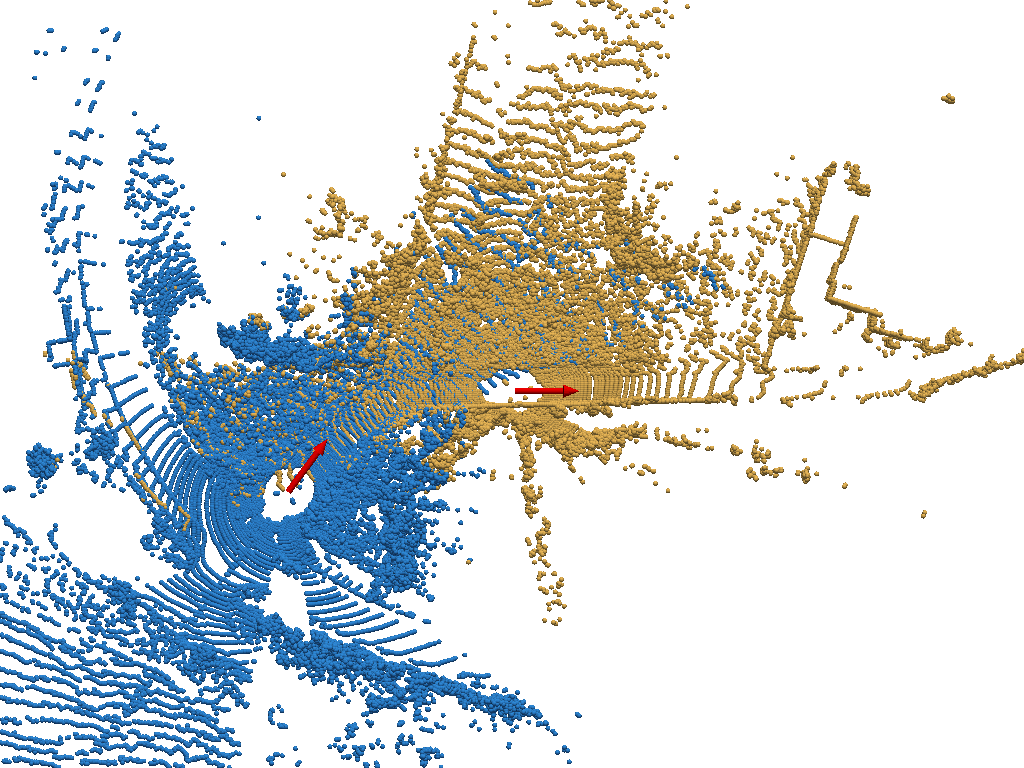}
        \end{minipage}
        \begin{minipage}[m]{0.24\linewidth}
        \centerline{$\RRE=0.4^\circ$}
        \centerline{$\RTE=0.03m$}
        \centering\includegraphics[width=0.99\linewidth]{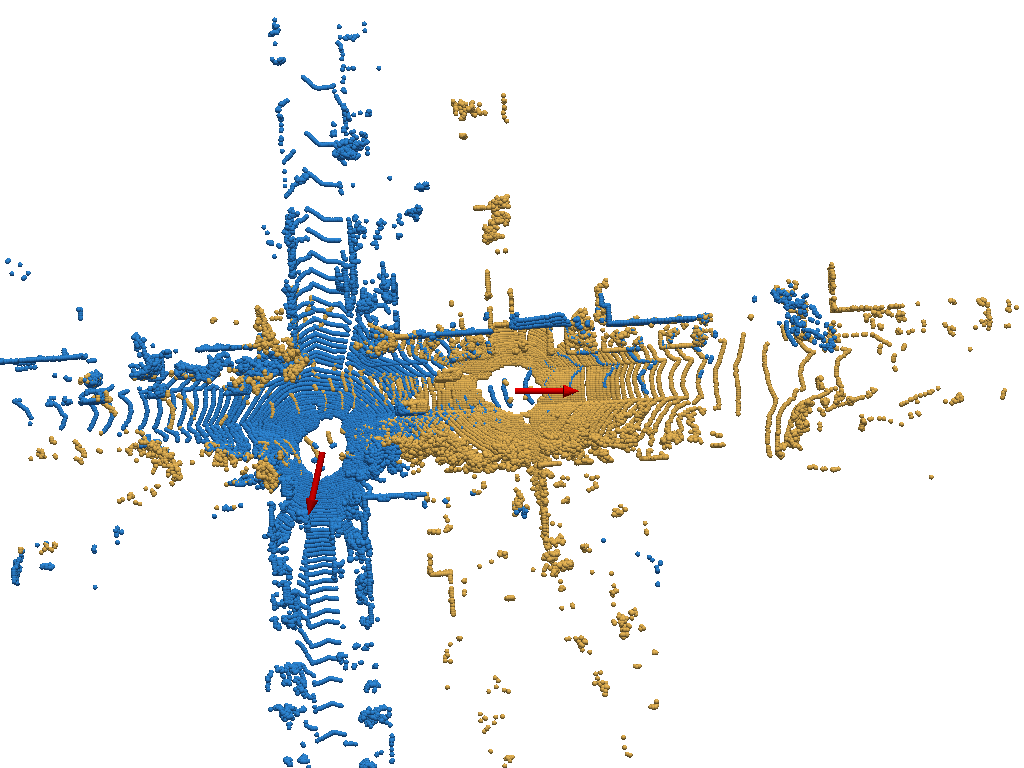}
        \end{minipage}
        \begin{minipage}[m]{0.24\linewidth}
        \centerline{$\RRE=1.9^\circ$}
        \centerline{$\RTE=0.6m$}
        \centering\includegraphics[width=0.99\linewidth]{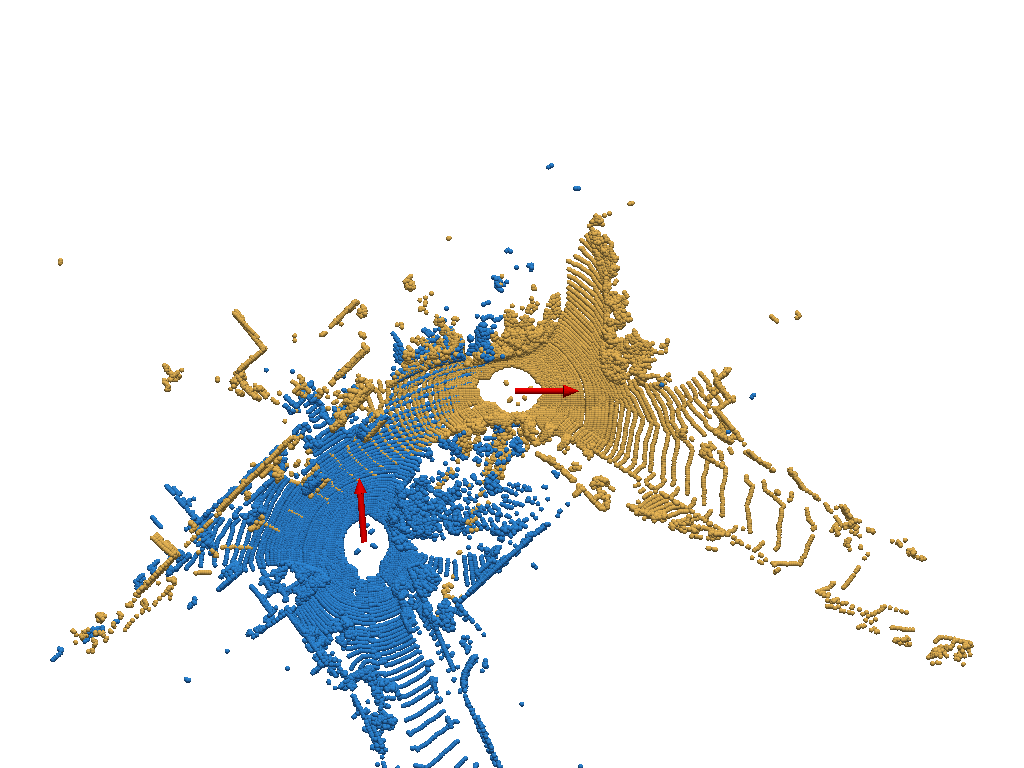}
        \end{minipage}
        \hrule
        \centerline{GCL}
        \begin{minipage}[m]{0.24\linewidth}
        \centerline{$\RRE=61^\circ$}
        \centerline{$\RTE=34m$}
        \centering\includegraphics[width=0.99\linewidth]{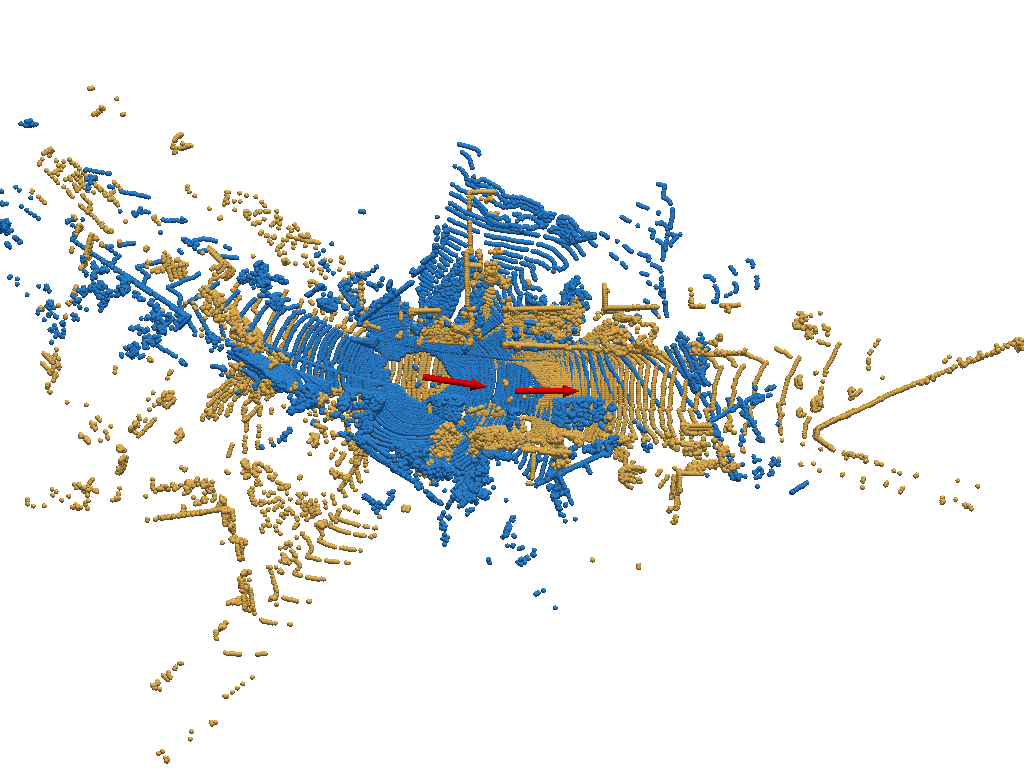}
        \end{minipage}
        \begin{minipage}[m]{0.24\linewidth}
        \centerline{$\RRE=1.4^\circ$}
        \centerline{$\RTE=1.5m$}
        \centering\includegraphics[width=0.99\linewidth]{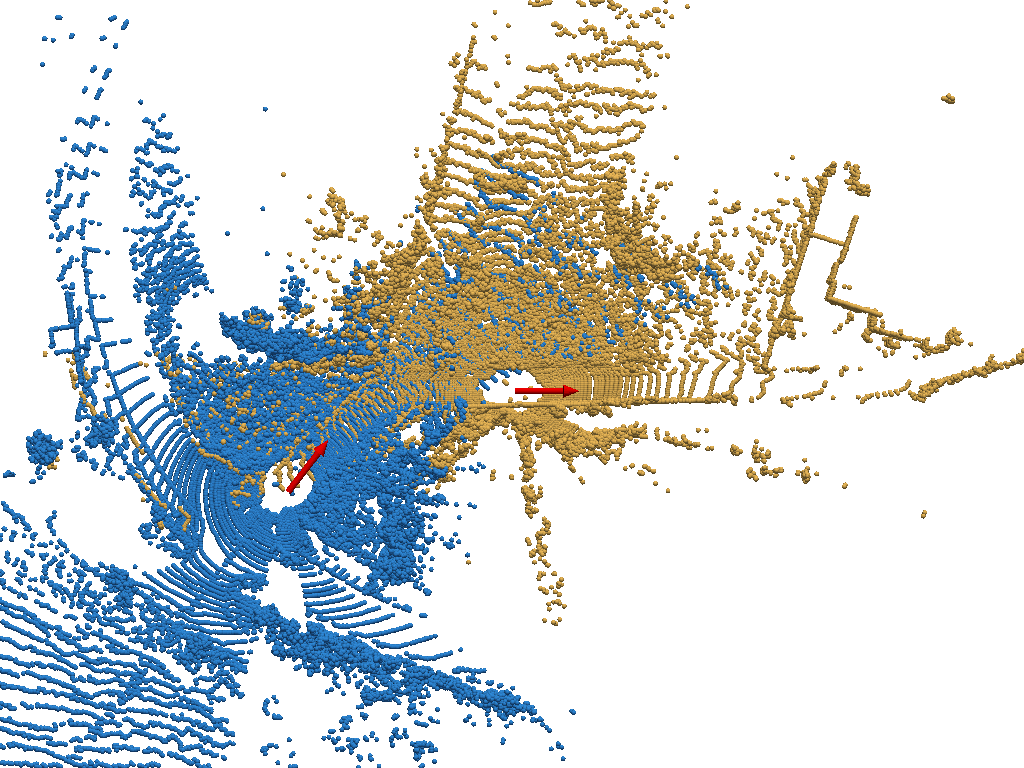}
        \end{minipage}
        \begin{minipage}[m]{0.24\linewidth}
        \centerline{$\RRE=102^\circ$}
        \centerline{$\RTE=31m$}
        \centering\includegraphics[width=0.99\linewidth]{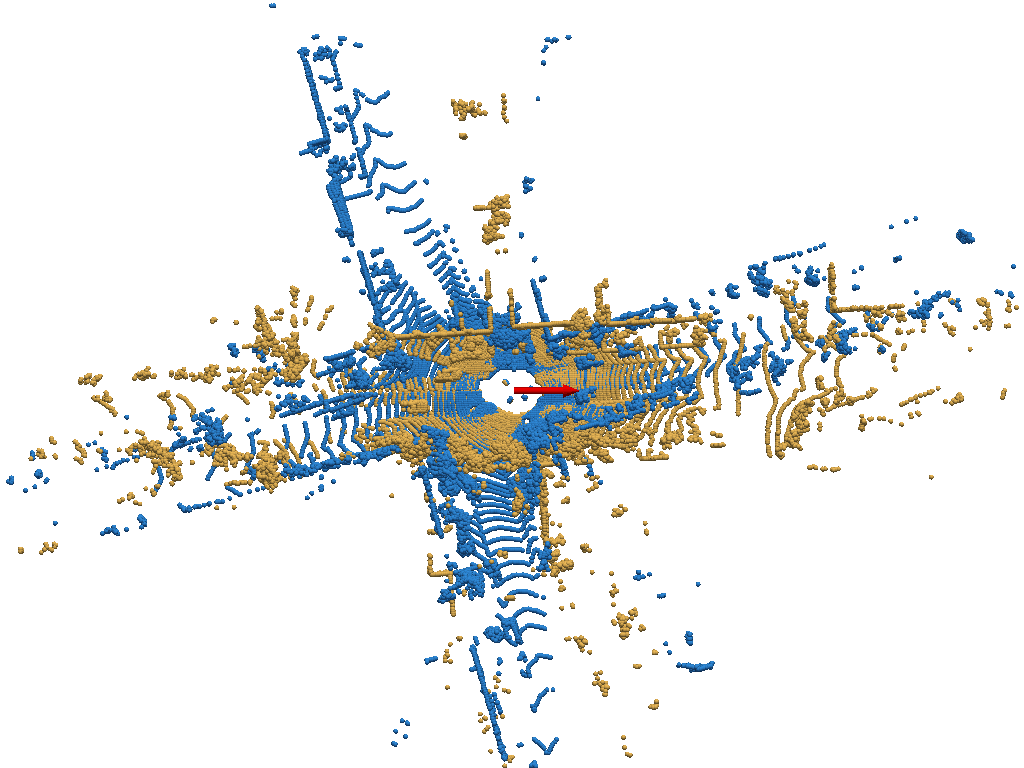}
        \end{minipage}
        \begin{minipage}[m]{0.24\linewidth}
        \centerline{$\RRE=97^\circ$}
        \centerline{$\RTE=32m$}
        \centering\includegraphics[width=0.99\linewidth]{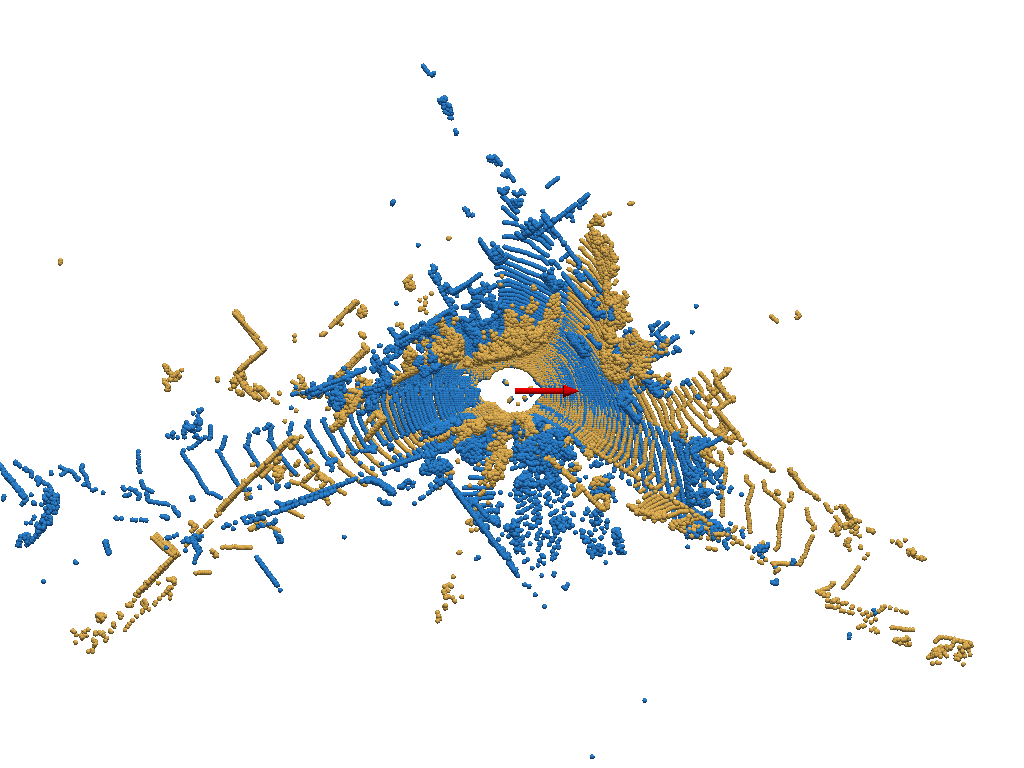}
        \end{minipage}
        \hrule
        \centerline{FCGF}
        \begin{minipage}[m]{0.24\linewidth}
        \centerline{$\RRE=52^\circ$}
        \centerline{$\RTE=50m$}
        \centering\includegraphics[width=0.99\linewidth]{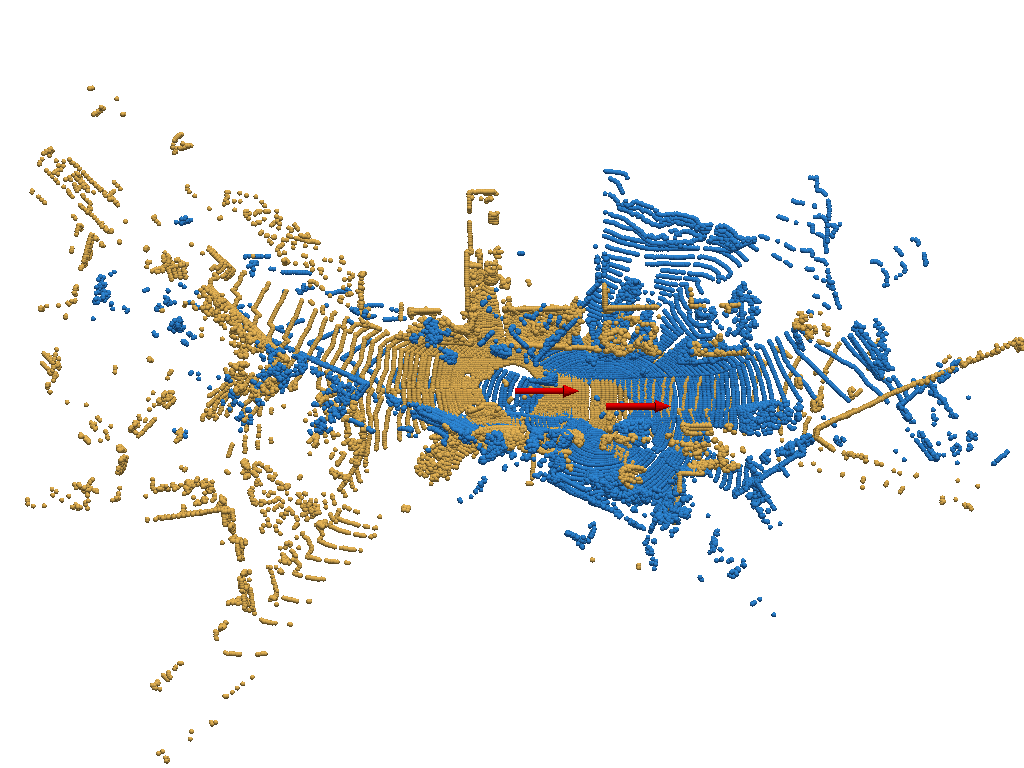}
        \end{minipage}
        \begin{minipage}[m]{0.24\linewidth}
        \centerline{$\RRE=54^\circ$}
        \centerline{$\RTE=38m$}
        \centering\includegraphics[width=0.99\linewidth]{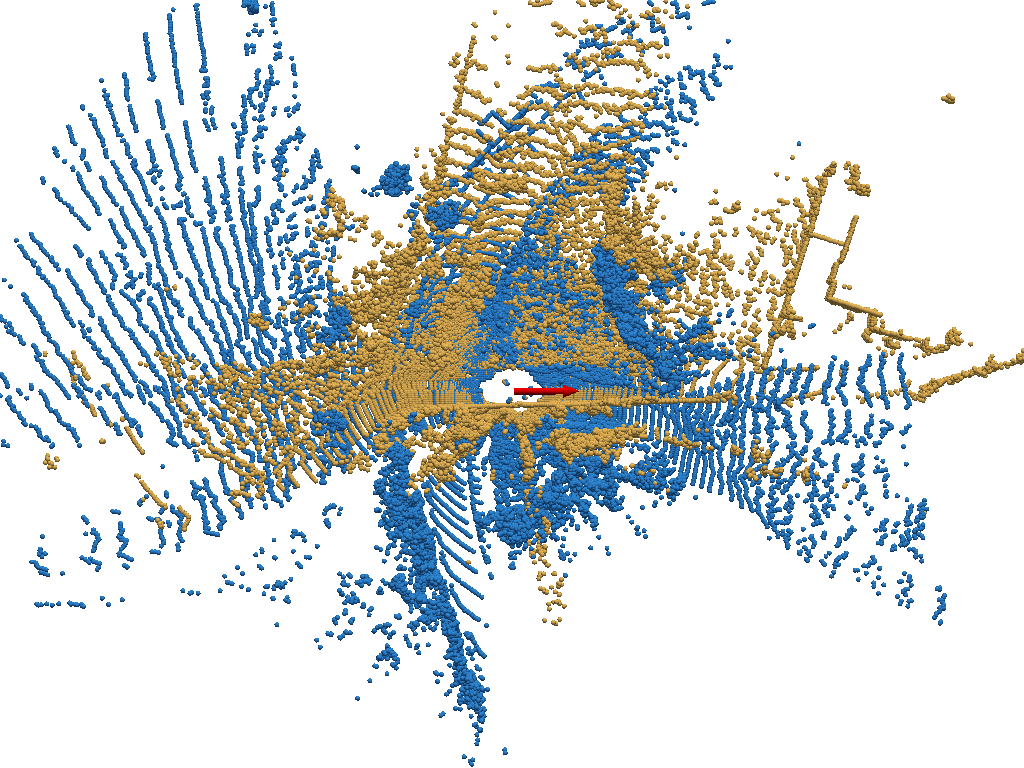}
        \end{minipage}
        \begin{minipage}[m]{0.24\linewidth}
        \centerline{$\RRE=102^\circ$}
        \centerline{$\RTE=31m$}
        \centering\includegraphics[width=0.99\linewidth]{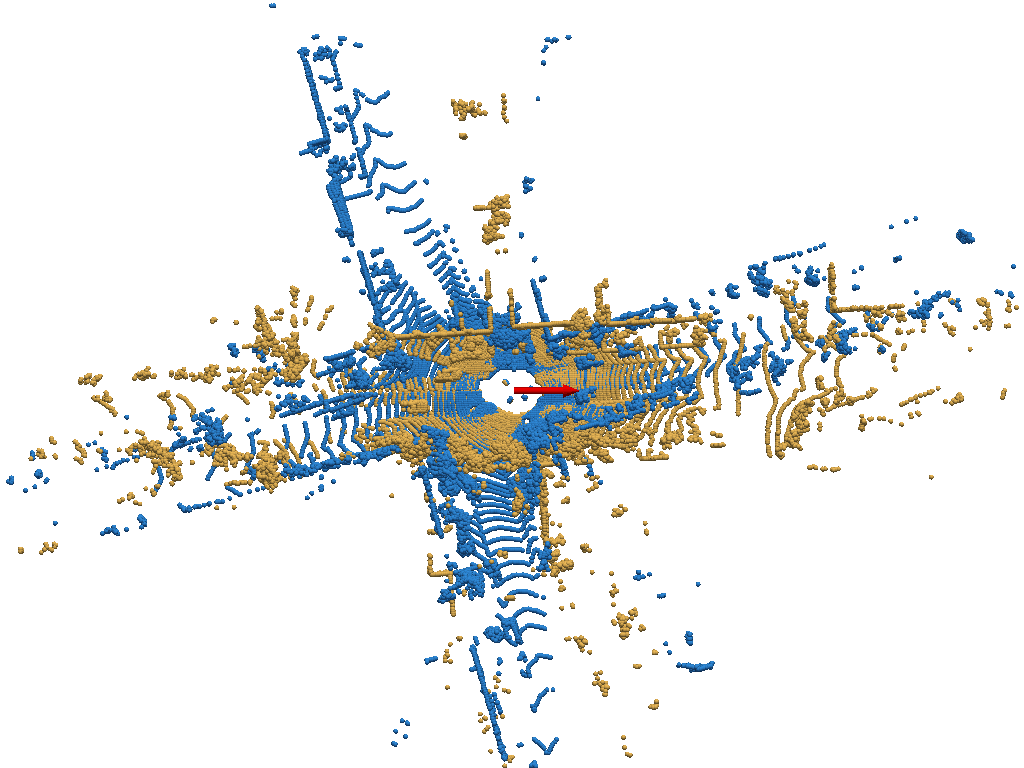}
        \end{minipage}
        \begin{minipage}[m]{0.24\linewidth}
        \centerline{$\RRE=99^\circ$}
        \centerline{$\RTE=37m$}
        \centering\includegraphics[width=0.99\linewidth]{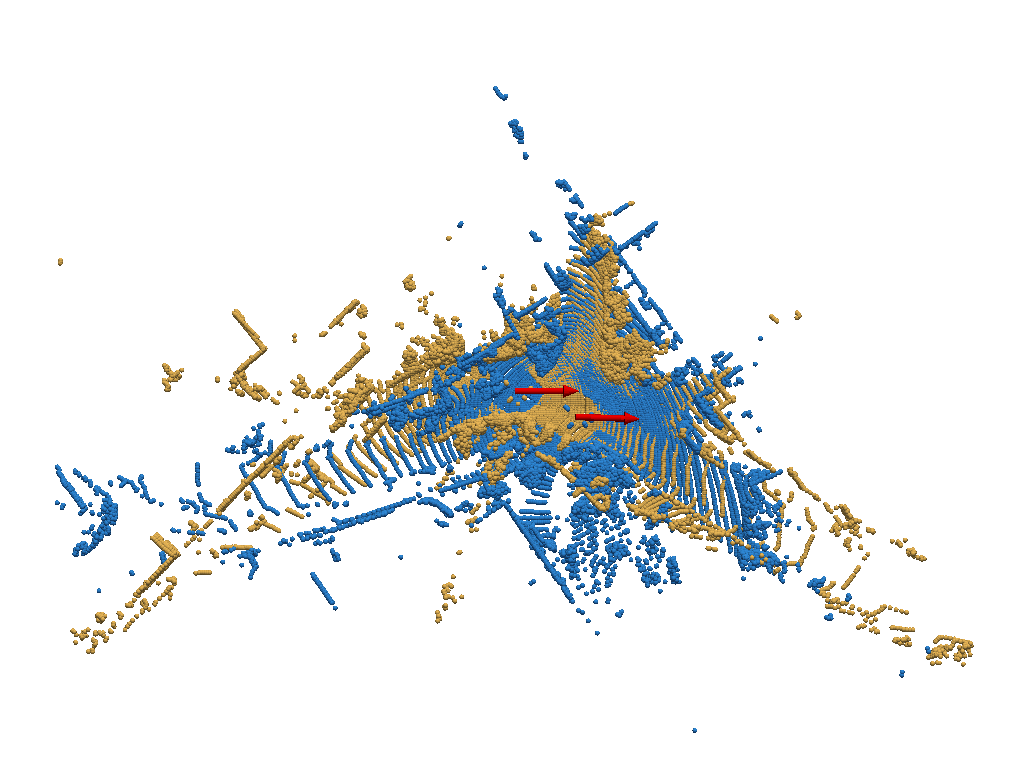}
        \end{minipage}
        \hrule
        \centerline{Predator}
        \begin{minipage}[m]{0.24\linewidth}
        \centerline{$\RRE=52^\circ$}
        \centerline{$\RTE=40m$}
        \centering\includegraphics[width=0.99\linewidth]{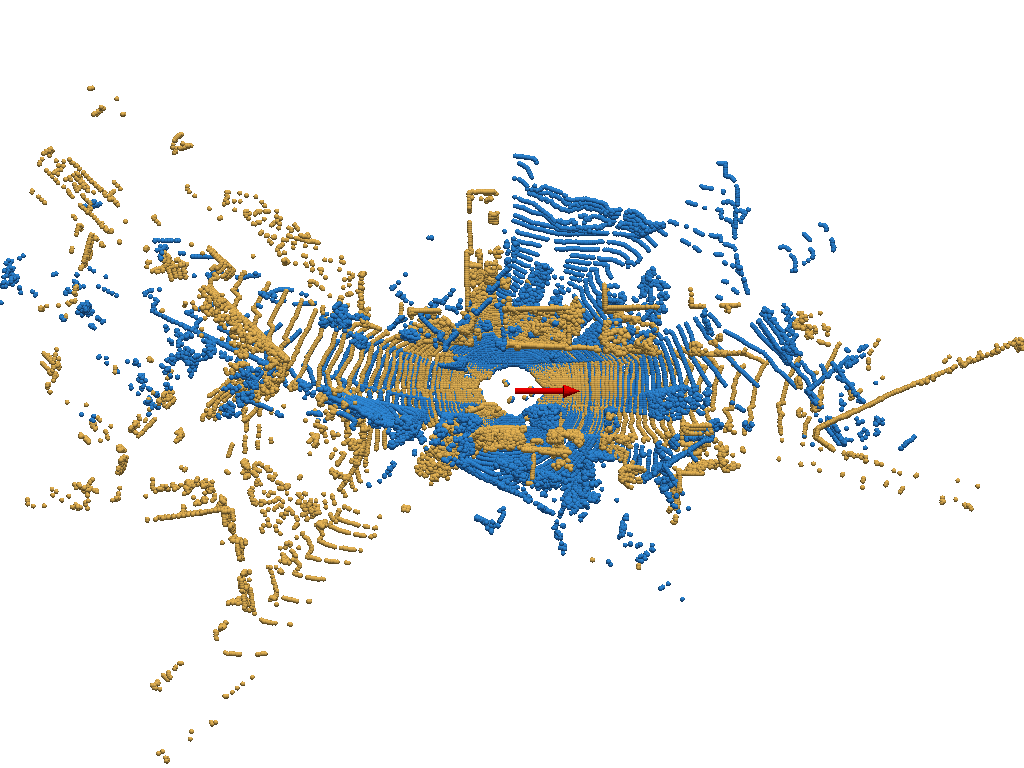}
        \end{minipage}
        \begin{minipage}[m]{0.24\linewidth}
        \centerline{$\RRE=54^\circ$}
        \centerline{$\RTE=38m$}
        \centering\includegraphics[width=0.99\linewidth]{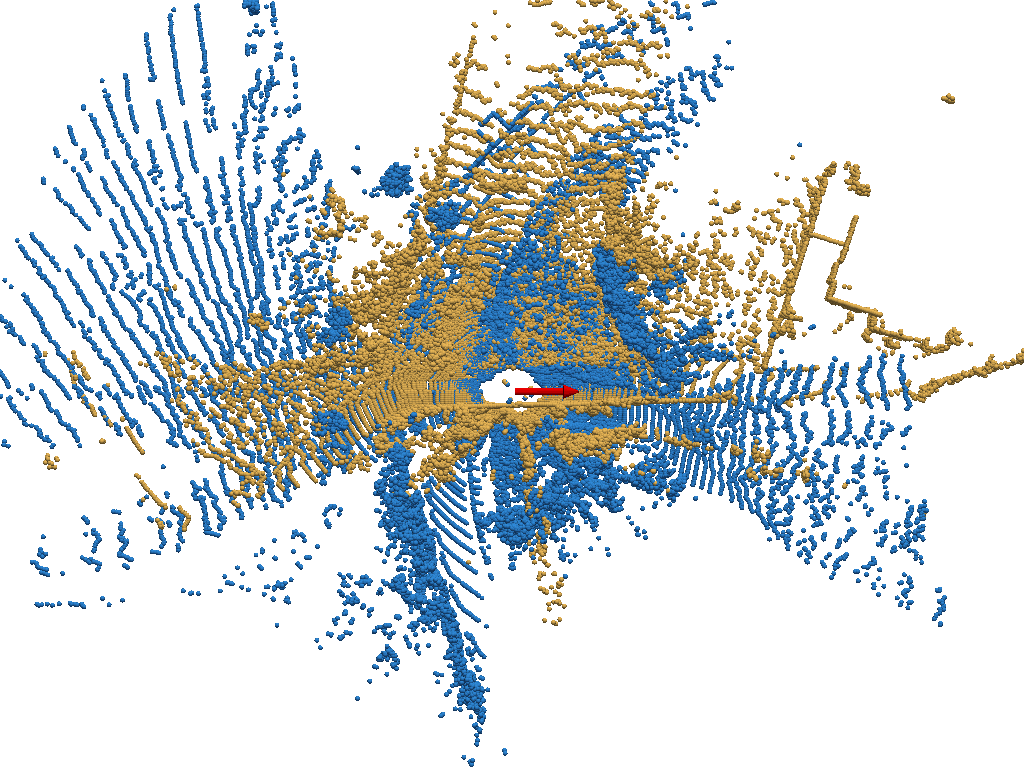}
        \end{minipage}
        \begin{minipage}[m]{0.24\linewidth}
        \centerline{$\RRE=102^\circ$}
        \centerline{$\RTE=31m$}
        \centering\includegraphics[width=0.99\linewidth]{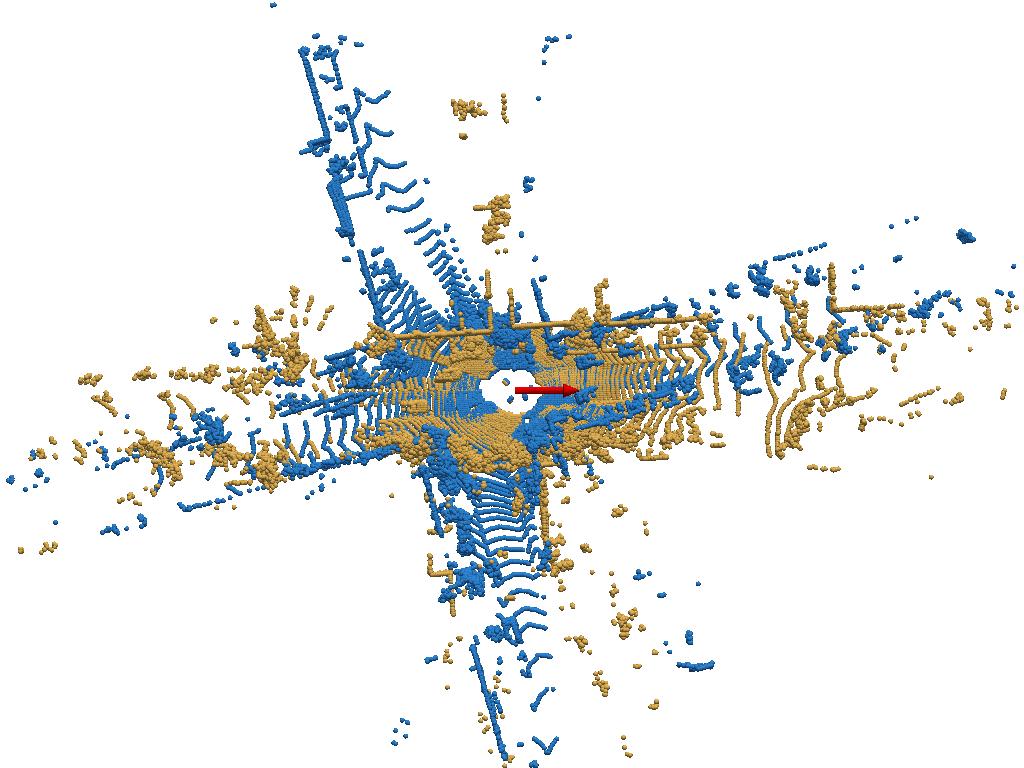}
        \end{minipage}
        \begin{minipage}[m]{0.24\linewidth}
        \centerline{$\RRE=97^\circ$}
        \centerline{$\RTE=32m$}
        \centering\includegraphics[width=0.99\linewidth]{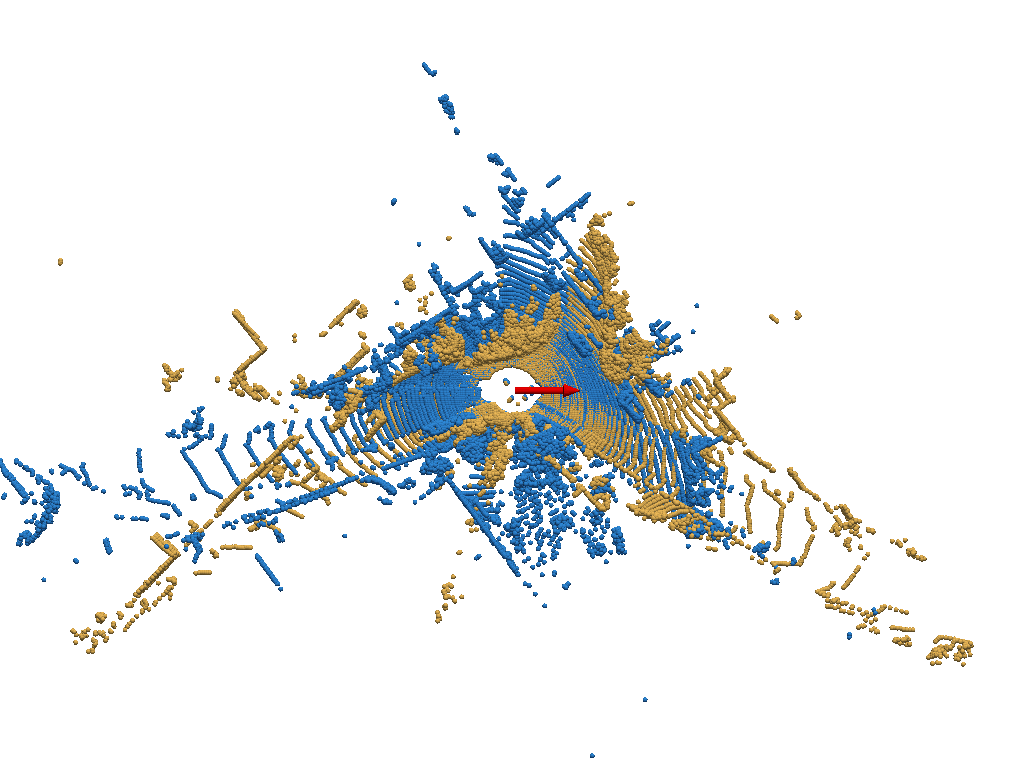}
        \end{minipage}
        \hrule
        \centerline{GeoTrans.}
        \begin{minipage}[m]{0.24\linewidth}
        \centerline{$\RRE=0.3^\circ$}
        \centerline{$\RTE=0.1m$}
        \centering\includegraphics[width=0.99\linewidth]{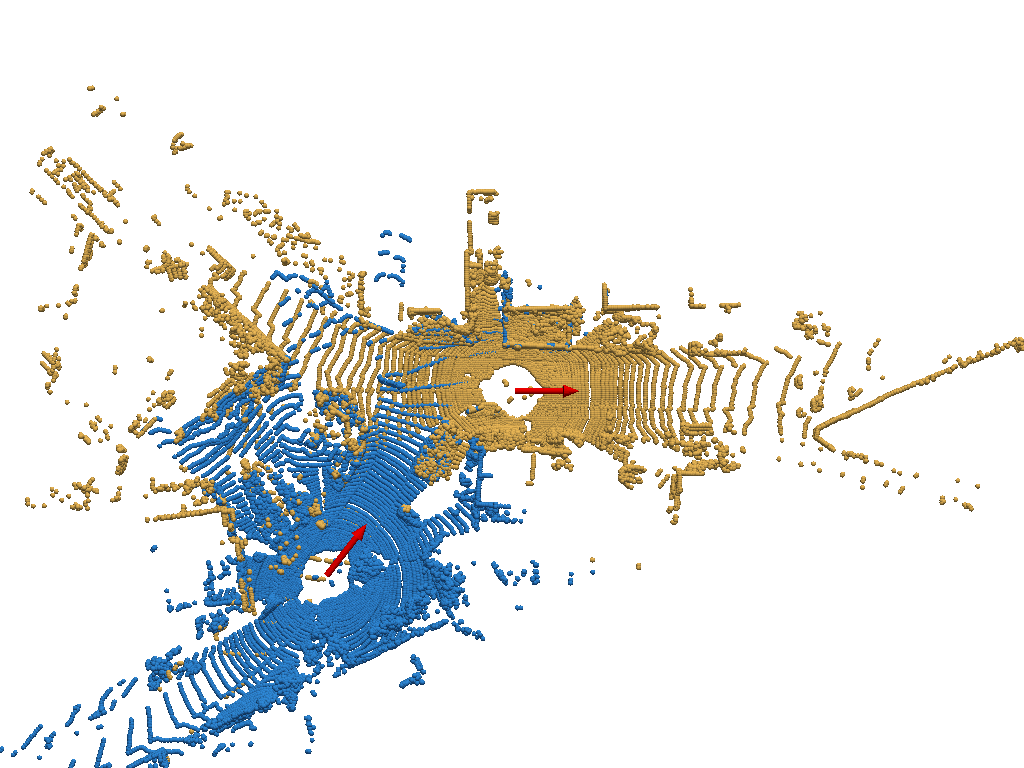}
        \end{minipage}
        \begin{minipage}[m]{0.24\linewidth}
        \centerline{$\RRE=94^\circ$}
        \centerline{$\RTE=48m$}
        \centering\includegraphics[width=0.99\linewidth]{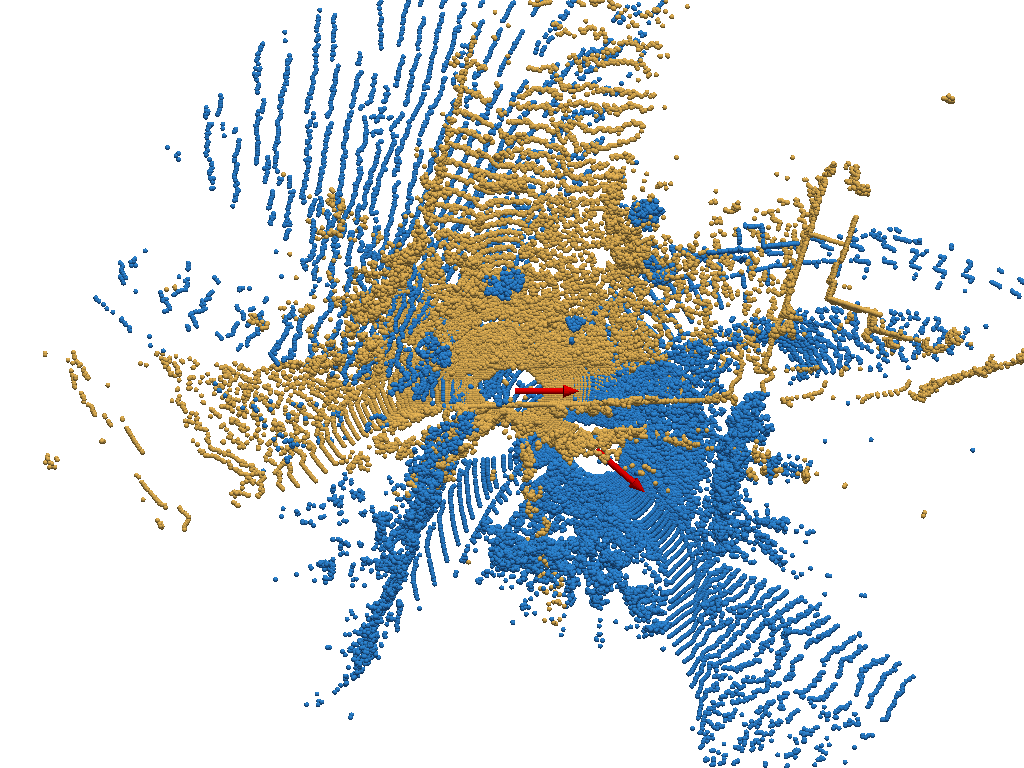}
        \end{minipage}
        \begin{minipage}[m]{0.24\linewidth}
        \centerline{$\RRE=114^\circ$}
        \centerline{$\RTE=53m$}
        \centering\includegraphics[width=0.99\linewidth]{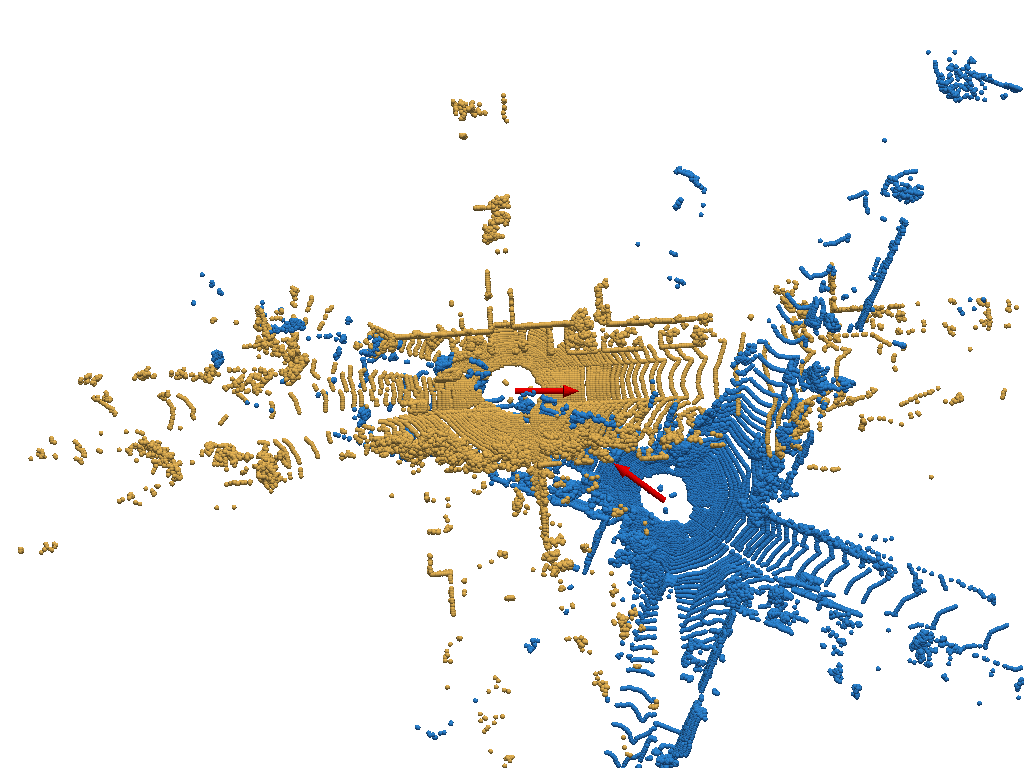}
        \end{minipage}
        \begin{minipage}[m]{0.24\linewidth}
        \centerline{$\RRE=40^\circ$}
        \centerline{$\RTE=48m$}
        \centering\includegraphics[width=0.99\linewidth]{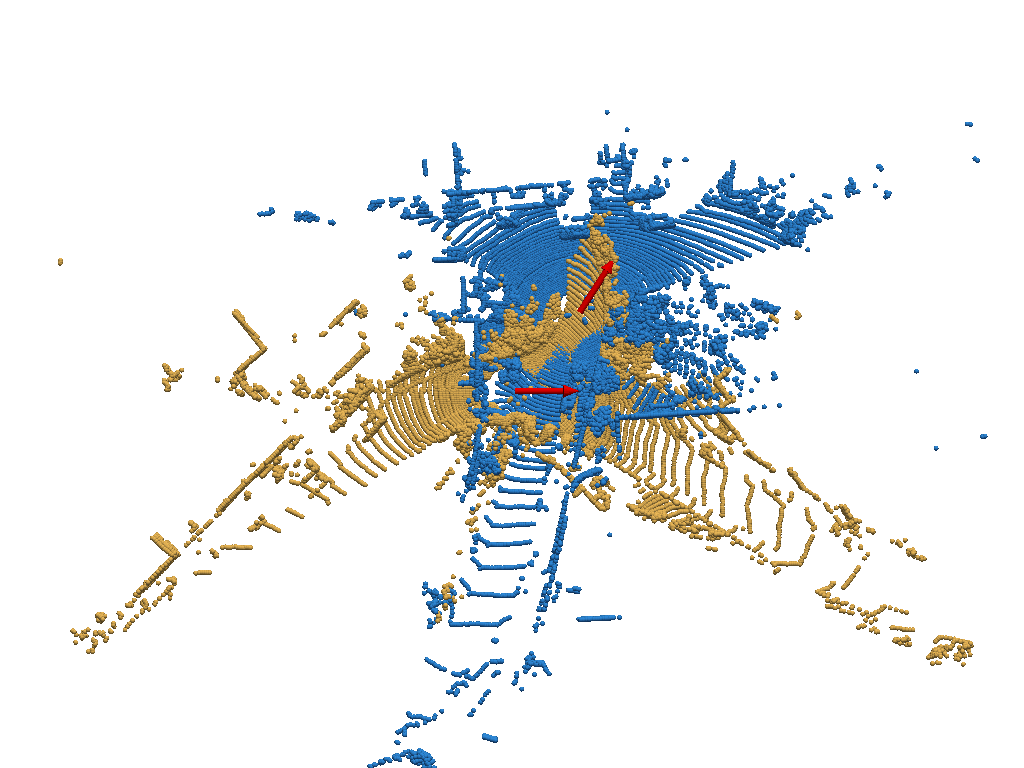}
        \end{minipage}
\caption{Qualitative Examples - Each column represents a different registration problem. Each sub-figure (except for the first row) depicts the  result of applying the estimated transformation to the source (blue) point cloud, to align it with the target (golden) point cloud.}\label{fig:qualitative_examples}
\end{figure}

In Fig. \ref{fig:qualitative_examples} we provide several qualitative examples extracted from the RotKITTI benchmark, presenting a comparison between UMERegRobust and the considered alternatives. The first row presents the ground truth (GT) transformation, indicating the relative rotation ($\theta$) in degrees and relative translation ($t$) in meters between the measurements. Additionally, the RRE and RTE for each method are provided. Each column represents a different scenario, while each row corresponds to a different method. Each sub-figure depicts the  result of applying the estimated transformation to the source (blue) point cloud, to align it with the target (golden) point cloud.

\end{document}